\pdfoutput=1

\documentclass[11pt]{article}

\usepackage[]{ACL2023}

\usepackage{times}
\usepackage{latexsym}
\usepackage{placeins}
\usepackage{booktabs} 

\usepackage{graphicx}
\usepackage{longtable}

\usepackage{graphicx} 
\usepackage{subcaption} 
\usepackage{svg}
\usepackage{tikz}
\usepackage{xcolor}
\usetikzlibrary{positioning,shapes,arrows,backgrounds,fit}
\usepackage{multirow}
\definecolor{boxbg}{HTML}{F8F9FA}
\definecolor{bordergray}{HTML}{E0E0E0}
\definecolor{textgray}{HTML}{4A4A4A}
\definecolor{smalltext}{HTML}{666666}
\definecolor{highlight}{HTML}{6366F1}
\definecolor{processbox}{HTML}{F8FAFC}
\definecolor{processborder}{HTML}{94A3B8}

\usepackage[T1]{fontenc}

\usepackage[utf8]{inputenc}

\usepackage{microtype}

\usepackage{inconsolata}

\usepackage{amsmath}
\usepackage{amssymb}
\usepackage{mathtools}
\usepackage{amsthm}

\usepackage[capitalize,noabbrev]{cleveref}

\theoremstyle{plain}

\theoremstyle{definition}

\theoremstyle{remark}

\usepackage[textsize=tiny]{todonotes}
\usepackage{enumitem}
\title{Sparse Autoencoder Features for Classifications and Transferability}

\author{
Jack Gallifant$^{1,2}$\textsuperscript{†}, 
Shan Chen$^{1,2,3}$\textsuperscript{†},  
Kuleen Sasse$^{4}$, 
Hugo Aerts$^{1,2,5}$, \\
\textbf{Thomas Hartvigsen$^{6}$, Danielle S. Bitterman$^{1,2,3}$\textsuperscript{§}} \\
\\
\textsuperscript{†}Co-first authors, \textsuperscript{§}Corresponding author: \texttt{dbitterman@bwh.harvard.edu} \\
\\
$^1$Harvard University, $^2$Mass General Brigham, $^3$Boston Children's Hospital, \\ 
$^4$Johns Hopkins University, $^5$Maastricht University, $^6$University of Virginia
}

\begin{document}
\maketitle

\begin{abstract}
Sparse Autoencoders (SAEs) provide potentials for uncovering structured, human-interpretable representations in Large Language Models (LLMs), making them a crucial tool for transparent and controllable AI systems. We systematically analyze SAE for interpretable feature extraction from LLMs in safety-critical classification tasks\footnote{Full repo: \url{https://github.com/shan23chen/MOSAIC}}. Our framework evaluates (1) model-layer selection and scaling properties, (2) SAE architectural configurations, including width and pooling strategies, and (3) the effect of binarizing continuous SAE activations. SAE-derived features achieve macro F1 > 0.8, outperforming hidden-state and BoW baselines while demonstrating cross-model transfer from Gemma 2 2B to 9B-IT models. These features generalize in a zero-shot manner to cross-lingual toxicity detection and visual classification tasks. Our analysis highlights the significant impact of pooling strategies and binarization thresholds, showing that binarization offers an efficient alternative to traditional feature selection while maintaining or improving performance. These findings establish new best practices for SAE-based interpretability and enable scalable, transparent deployment of LLMs in real-world applications.
\end{abstract}

\section{Introduction}
\label{sec:intro}
\begin{figure}[!htp]
    \centering
    \includegraphics[width=\linewidth]{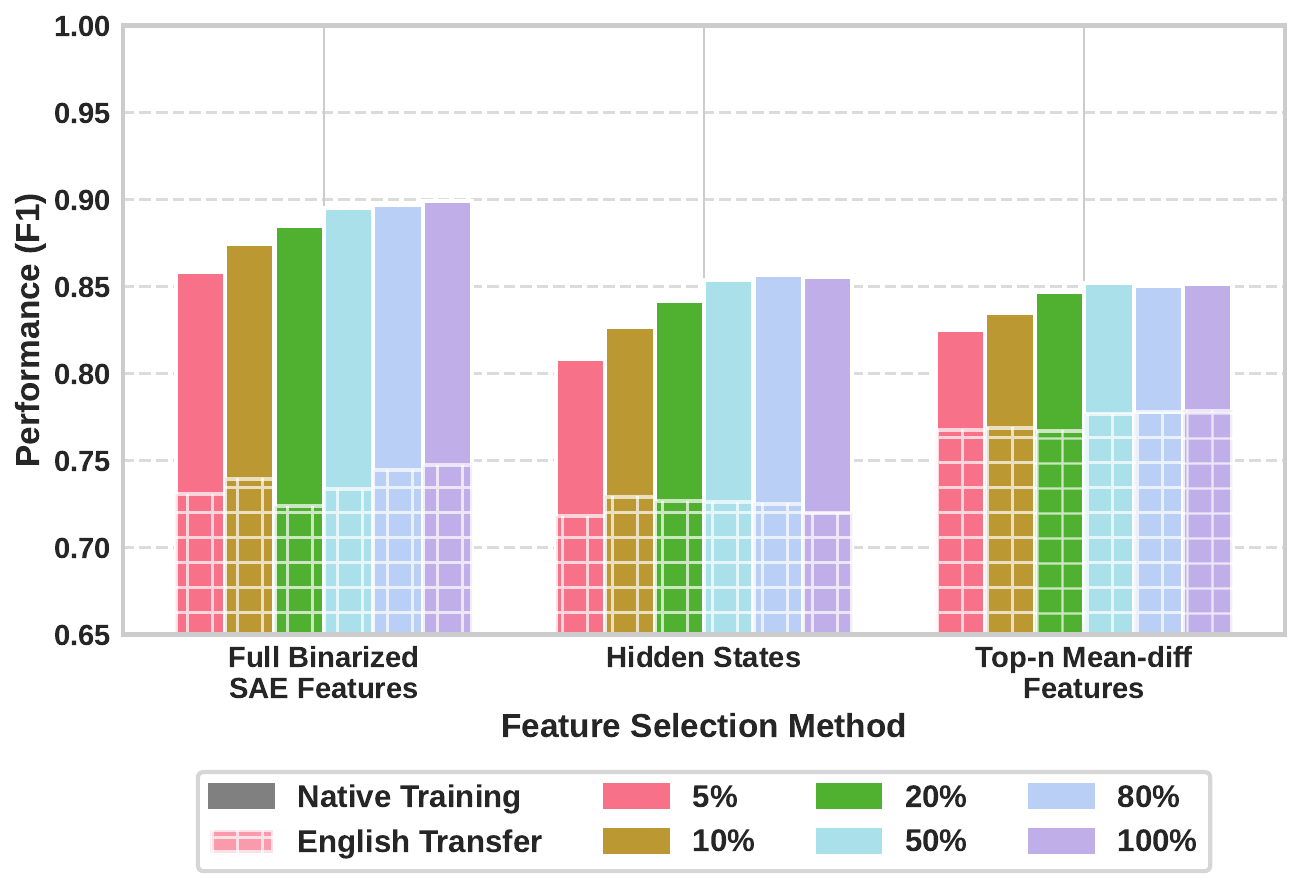}
    \caption{Multilingual performance comparison across three feature selection methods under varying training data sampling rates. Solid bars represent models trained on native language data, while hatched bars show performance with English transfer learning. Binarized SAE features demonstrate robustness across different training data constraints.}
    \label{fig:sampling-main}
\end{figure}

Large language models (LLMs) have transformed natural language processing (NLP), demonstrating impressive performance on diverse tasks and languages, even in knowledge-intensive and safety-sensitive scenarios \cite{hendrycks2023overview, ngo2022alignment, cammarata2021curve}.  However, the internal decision-making processes of LLMs remain largely opaque \cite{cammarata2021curve}, raising concerns about trustworthiness and oversight, especially given the potential for deceptive or unintended behaviors. Mechanistic interpretability (MI), the study of the internal processes and representations that drive a model's outputs, offers a promising approach to address this challenge \cite{elhage2022toy, wang2022interpretability}. However, despite its potential, applying MI to real-world tasks presents significant challenges.

\begin{figure*}[ht]
    \centering
    \includegraphics[width=\linewidth]{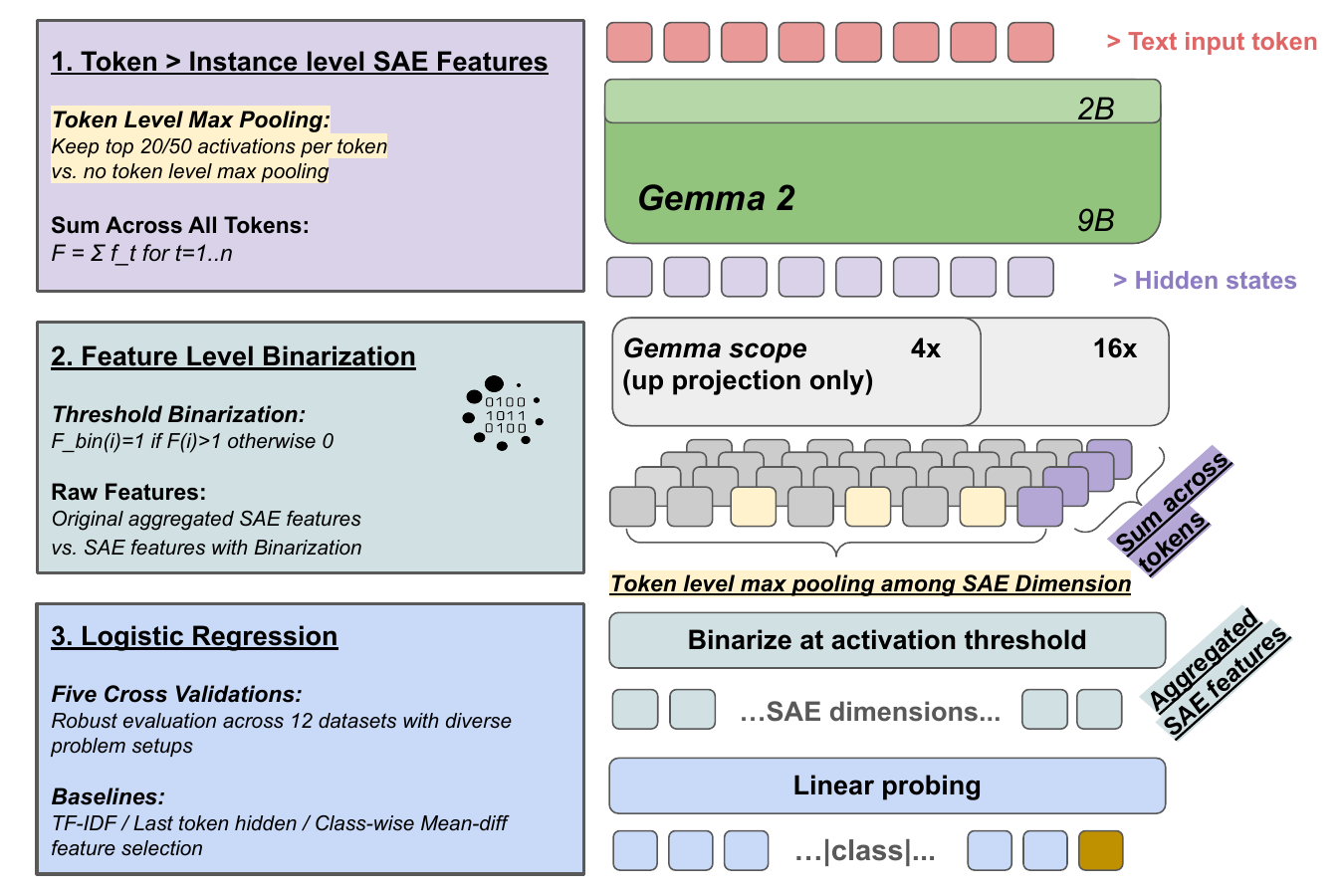}
    \caption{Diagram explaining our approaches to evaluating token-level pooling and aggregation of SAE features.}
    \label{fig:workflow}
\end{figure*}
\raggedbottom

Sparse Autoencoders (SAEs) have recently emerged as a promising technique within MI for understanding LLMs. SAEs generally work by learning a compressed, sparse representation of the LLM's internal activations.  This is achieved by up-projecting the dense hidden state of the LLM to a sparser, ideally monosemantic, representation \cite{bricken2023towards, gao2024scalingevaluatingsparseautoencoders}. Identifying semantically meaningful features within LLMs using SAEs allows for deploying these features into explainable classification pipelines. This has the potential to boost performance and detect harmful biases or spurious correlations before they manifest in downstream tasks \cite{anthropic2024features}.  The ability to employ SAE features for classification across diverse settings, ranging from toxicity detection to user intent, offers a scalable form of "model insight" \cite{bowman2022measuringprogressscalableoversight}, which is crucial for building trust, safety, and accountability in high-stakes domains like medicine and law \cite{abdulaal2024xrayworth15features}.

Despite the promise of SAEs for MI, surprisingly few systematic studies have provided practical guidance on their use for classification. While promising results have been reported across various tasks \cite{anthropic2024features, sae_probing, sae_features_llava}, inconsistent experimental protocols, a lack of standardized benchmarks, and limited exploration of key architectural decisions hinder comparability and the development of best practices. Although tools like Transformer Lens \cite{nanda2022transformerlens} and SAE Lens \cite{bloom2024saetrainingcodebase} have improved standardization in sampling activations, critical questions about optimal configurations for diverse tasks, particularly in multilingual and multimodal settings, remain unanswered. This makes it challenging to establish the robustness and generalizability of SAE-based classification approaches.

This work directly addresses these limitations by providing a comprehensive and systematic investigation of SAE-based classification for LLMs. We introduce a reproducible pipeline for large-scale activation extraction and classification, enabling robust and generalizable conclusions. Specifically, we explore critical methodological choices, evaluate performance across diverse datasets and tasks, and investigate the potential for SAEs to facilitate model introspection and oversight (Figure \ref{fig:workflow}).

\paragraph{Summary of Contributions}
\begin{enumerate}[itemsep=-1.7pt,topsep=1.5pt]
    \item \textit{Systematic Classification Benchmarks (Section~\ref{sec:pooling}, Part~1 ):} We introduce a robust methodology to evaluate and select SAE-based features in safety-critical classification tasks and show superior performance overall. 
    
    \item \textit{Multilingual Transfer Analysis (Section~\ref{sec:multilingual}, Part~2):} We analyze the cross-lingual transferability of SAE features in multilingual toxicity detection and show SAE features outperform everything in-domain and demonstrate potential on cross-lingual feature generalization.
    
    \item \textit{Behavioral Analysis and Model Oversight (Section~\ref{sec:action}, Part~3):} We extend SAE-based features to model introspection tasks, investigating whether LLMs can predict their own correctness and that of larger models, showing the potential of scalable model oversight.
    

\end{enumerate}

\section{Related Work}
\label{sec:related}
\subsection{Interpretable Feature Extraction}

MI has evolved from neuron-level analysis to sophisticated feature extraction frameworks \cite{olah2020zoom, rajamanoharan2024improving}. Early approaches targeting individual neurons encountered fundamental limitations due to polysemanticity, where activation patterns span multiple, often unrelated concepts \cite{bolukbasi2021interpretability, elhage2022toy}. While techniques like activation patching \cite{meng2022locating} and attribution patching \cite{syed2023attribution} offered insights into component-level contributions, they highlighted the need for more comprehensive representational frameworks.

SAEs address these limitations by providing more interpretable feature sets \cite{bricken2023towards, cunningham2023sparse}. Recent scaling efforts have demonstrated SAE viability across LLMs from Claude 3 Sonnet \cite{templeton2024scaling} to GPT-4 \cite{gao2024scaling} with extensions to multimodal architectures like CLIP \cite{bhalla2024interpreting}. Although these studies have revealed interpretable feature dimensions and computational circuits \cite{marks2024sparse, zhao2024steeringknowledgeselectionbehaviours}, they focus mainly on descriptive feature discovery rather than systematic evaluation of their downstream applications. Our work bridges this gap by providing standardized evaluation frameworks for SAE-based classification and cross-modal transfer, establishing quantitative metrics and methods for feature utility across diverse tasks.

\subsection{SAE-Based Classification and its Limitations}
Reports have demonstrated that SAE-derived features can outperform traditional hidden-state probing for classification, particularly in scenarios with noisy or limited data with closed datasets \cite{anthropic2024features} or simplified tasks \cite{sae_probing}. However, more recent studies, such as \citet{wu2025axbenchsteeringllmssimple}, suggest that SAEs may not be superior, particularly for model steering (instead of classification). These seemingly conflicting results highlight a critical gap in the current understanding of SAE-based classification: a lack of systematic exploration of how hyperparameters, feature aggregation strategies, and other methodological choices impact performance. 

Existing evaluations often focus on narrow settings, making it unclear whether discrepancies arise from task differences, dataset choices, or specific configurations. This work addresses this gap by systematically evaluating SAE-based classification. We examine key hyperparameters and methodological choices like feature pooling, layer selection, and SAE width across diverse datasets and tasks, ensuring a fair comparison with established baselines.

\section{Preliminaries}
\label{sec:preliminaries}
\paragraph{Experimental Setup Rationale:}
Our primary goal is to evaluate pre-trained SAE features for interpretable, zero-shot classification tasks. Accordingly, we selected the Gemma 2 SAE suite as it was the only publicly available family offering matched model backbones (2B, 9B, 9B-IT) with identical training settings and systematic layer and width pairings. We compare against two standard interpretable baselines: linear probes on hidden-state activations and TF-IDF on a bag-of-words representation. We deliberately exclude fine-tuned models, as they operate under a different, less-interpretable paradigm and fall outside our zero-shot evaluation scope. The TF-IDF baseline serves as a strong, classic non-neural benchmark for interpretability and performance.

\paragraph{Notation and Setup:}
Let $M$ be a pretrained LLM with hidden dimension $d$. When $M$ processes an input sequence of tokens of length $n$, it produces hidden representations $\{\mathbf{h}_1, \mathbf{h}_2, \ldots, \mathbf{h}_n\}$ for each layer, where each $\mathbf{h}_t \in \mathbb{R}^d$. We consider three versions of \texttt{Gemma 2} models \cite{gemmateam2024gemma2improvingopen} in this work, the \textbf{2B},  \textbf{9B} and instruction-tuned variant, \textbf{9B-IT}.

\paragraph{SAE-Based Activation Extraction:}
We use \textbf{pretrained} SAEs provided by \texttt{Gemma Scope} \cite{lieberum2024gemma}, choosing the SAE with $L_0$ loss closest to 100. We extract each token’s residual stream activations from layers that have been instrumented with the \texttt{SAELens} \cite{bloom2024saetrainingcodebase} tool. Specifically for the 2B model, we extract SAE features from layers 5, 12, 19 (early, middle, late) where 9B \& 9B-IT models with layers 9, 20, and 31 from the residual stream.

Each SAE has a designated \emph{width} (i.e., number of feature directions). We evaluate \textbf{16K} and \textbf{65K} widths for the 2B model, and \textbf{16K} and \textbf{131K} for 9B and 9B-IT \footnote{we choose 131k for 9B and 65k for 2B models due to their same expansion ratio to original model hidden states}, following the pretrained SAEs made available in \texttt{Gemma Scope} \cite{lieberum2024gemma}. \textbf{Note}: we do \textit{not} train any SAEs ourselves; our workflow involves only extracting the hidden states and the corresponding \emph{pretrained} SAE activations.

\paragraph{Pooling and Binarization}
Since SAEs generate token-level feature activations, an essential step in classification is aggregating these activations into a fixed-size sequence representation. Without pooling, the model lacks a structured way to combine token-level representations. Previous NLP works have explored various pooling strategies for feature aggregation in neural representations \cite{shen2018baseline}. However, it remains unclear which pooling method is most effective for LLMs' SAE features. We systematically evaluate different pooling approaches (displayed in \ref{fig:workflow}, considering (1) \emph{Top-$N$ feature selection per token} \footnote{Token-level top-$N$ where n=0 indicates the absence of max pooling. \cite{karvonen2025saebenchcomprehensivebenchmarksparse}} and (2) \emph{summation-based aggregation}\footnote{this approach is also adopted by parallel research \cite{brinkmann2025large}.} which collapses token-level activations into a single sequence vector:

\begin{equation}
    \mathbf{F} = \sum_{t=1}^n \mathbf{f}_t,
\end{equation}

where $\mathbf{f}_t \in \mathbb{R}^m$ is the SAE feature vector of dimension $m$ for token $t$. The summation method aggregates all token activations, while top-n selects the strongest activations per token. Further details are provided in \ref{app:pooling_methods}.

Beyond pooling, we investigate \emph{binarization} to enhance interpretability and efficiency. This transformation converts $\mathbf{F}$ into a binary vector $\mathbf{F}_{\text{bin}}$, activating only the dimensions that exceed a threshold:

\begin{equation}
    \mathbf{F}_{\text{bin}}[i] = 
    \begin{cases}
      1, & \text{if } \mathbf{F}[i] > 1, \\
      0, & \text{otherwise}.
    \end{cases}
\end{equation}

Binarization provides multiple advantages: (1) it produces compact, memory-efficient representations, (2) it acts as a non-linear activation akin to ReLU \cite{agarap2019deeplearningusingrectified}, and (3) it serves as an implicit feature selection mechanism, highlighting only the most salient SAE activations. By thresholding weaker activations, this approach enhances the robustness and interpretability of extracted features in downstream classification tasks.

\paragraph{Classification with Logistic Regression:}
To measure how informative these SAE-derived features are for various tasks, we train a \emph{logistic regression} (LR) classifier. In all experiments, LR models are evaluated using \textbf{5-fold cross-validation}. This is the only learned component of our pipeline;
\paragraph{Baselines:} We compare against:
\begin{itemize}[itemsep=-1.7pt,topsep=1.5pt]
    \item \textbf{TF-IDF}: Classic bag-of-words variation without neural representations \cite{sparck_jones_1972}.
    \item \textbf{Hidden State}: Like prior studies \cite{features_as_classifiers}, we did compare to \emph{last-token} hidden state probing as well. 
\end{itemize}

\paragraph{Code and Reproducibility:}
All code for data loading, activation extraction, pooling, detailed hyper-parameters and classification results is provided in a public repository. A simple YAML configuration file controls model scale, layer indices, SAE width, and \texttt{huggingface} dataset paths, enabling reproducible workflows with Apache 2 license. All our experiments are conducted on three Nvidia A6000 GPUs with CUDA version 12.4.

\section{Classification Tasks, Multimodal Transfer, and Hyperparameter Analysis}
\label{sec:pooling}
Here, we investigate best practices for using \emph{GemmaScope} SAE features in classification tasks across model scale, SAE width, layer depth, pooling strategies, and binarization. We also briefly touch upon the cross-modal applicability of text-trained SAE features to a \emph{PaliGemma 2} vision-language model.

\paragraph{Datasets:}
We targeted scalable, safety-relevant \emph{binary} classification tasks—jailbreak detection, harmful-prompt screening, and multilingual toxicity—to stress-test generality while keeping evaluation simple and comparable. Concretely, we selected publicly available datasets drawn from MTEB and other widely used classification corpora to ensure reproducibility and sufficient scale \cite{muennighoff2023mtebmassivetextembedding}. We prioritized (i) clear binary labels, (ii) coverage across multiple languages, and (iii) permissive licensing. At the time of experimentation, the pool of multilingual binary datasets was limited, so we focused on these three tasks; broadening the task set is an important direction for future work. Detailed dataset characteristics are in Appendix~\ref{app:model_info}.

\begin{figure}[t]
    \centering
    \begin{subfigure}[b]{\linewidth}
        \centering
        \includegraphics[width=\linewidth]{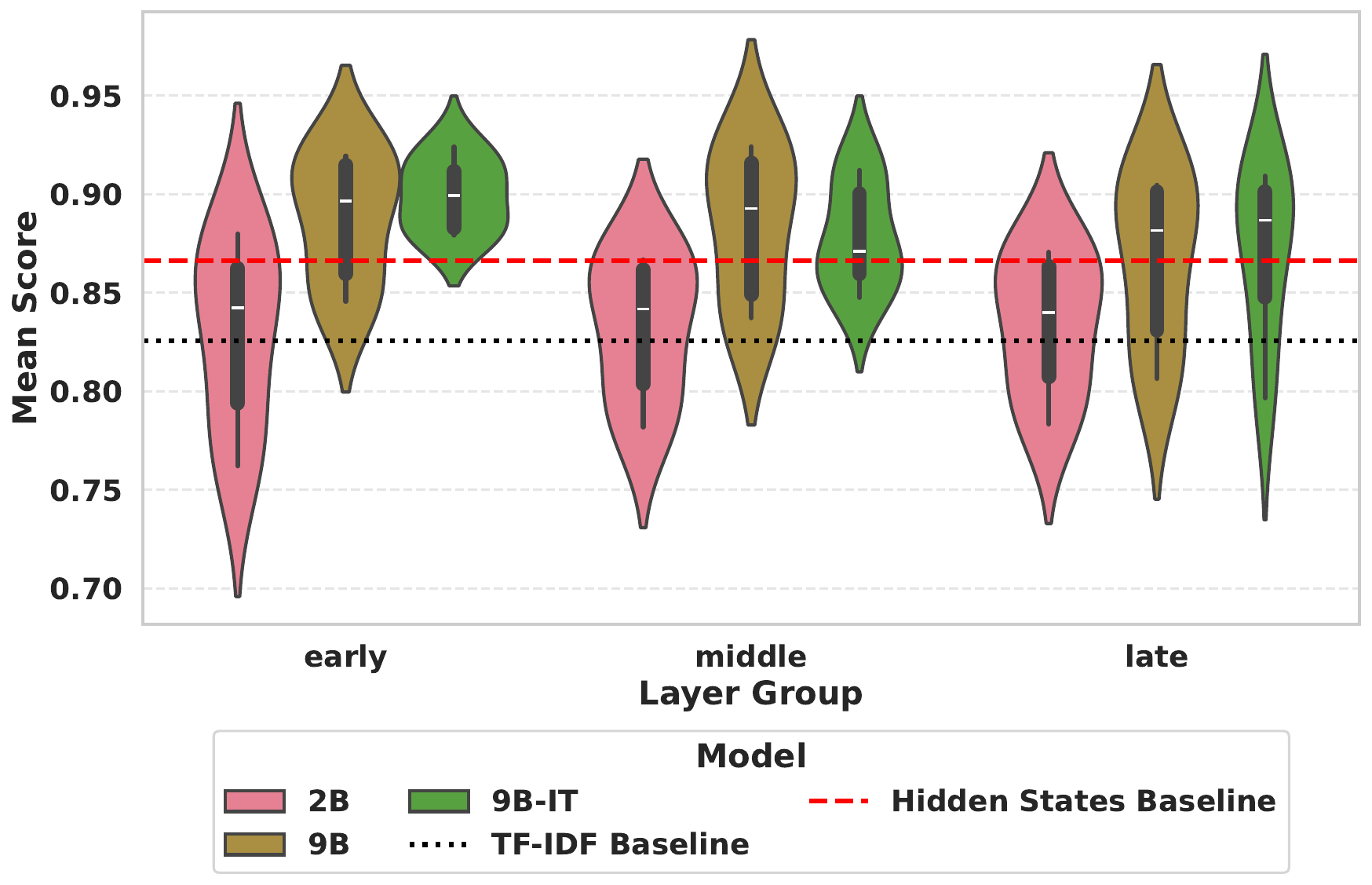}
        \caption{Layer-wise classification performance for each model scale. The dotted black line indicates a TF-IDF baseline, while the red dashed line indicates a last token hidden-state probe baseline. SAE-based methods (colored violin plots) often surpass these baselines, with middle-layer SAE features typically achieving the highest scores.}
        \label{fig:pooling_a}
    \end{subfigure}
    
    \begin{subfigure}[b]{\linewidth}
        \centering
        \includegraphics[width=\linewidth]{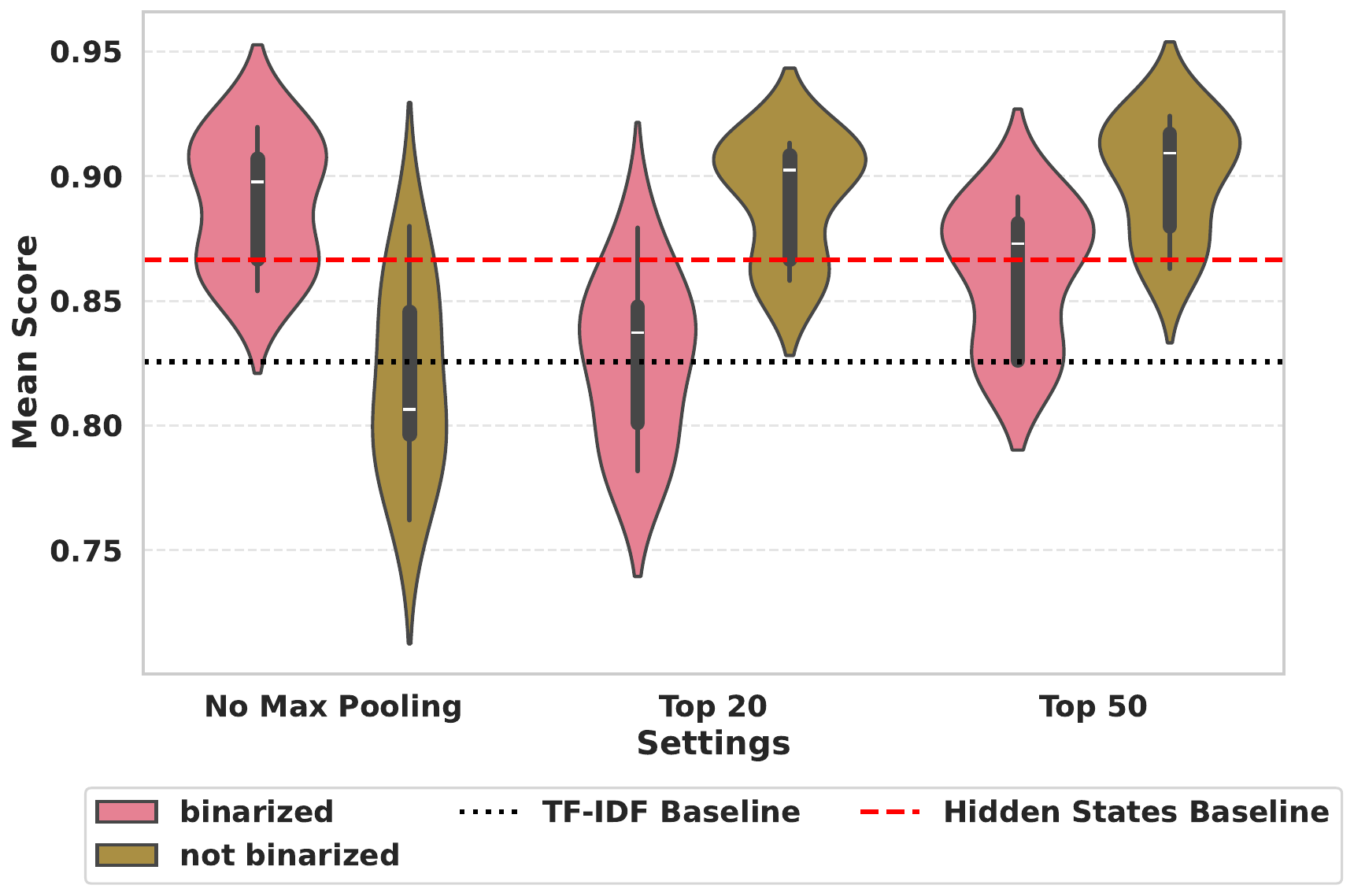}
        \caption{Token level top-$N$ vs.\ full binarized features. Token level top-$N$ improves with larger values of $N$, and binarization can worsen this performance. However, binarization of all tokenwise activations reached the best performance of Token level top-$N$ whilst removing the need to compute top-$N$ values, which would be important as $N$ scales, offering a more efficient alternative.}
        \label{fig:pooling_b}
    \end{subfigure}
    \caption{Analysis of model performance across different layers and pooling strategies. A strong baseline is established by averaging the optimal performance per task across the hidden states across three models.}
    \label{fig:pooling}
\end{figure}

\subsection{Impact of Layer Depth and Model Scale}

We evaluate \texttt{gemma-2-2b}, \texttt{9b}, and \texttt{9b-it}, using their early, middle, and late layers, with SAE widths of 16K/65K for \texttt{gemma-2-2b} and 16K/131K for \texttt{gemma-2-9b} and \texttt{9b-it}, using different pooling strategies.

We extract token-level SAE features and train LR classifiers, comparing the results to TF-IDF and final-layer hidden-state baselines \footnote{we did not benchmark against mean-diff here because that required task to be binary classification}. Figure~\ref{fig:pooling}(a) depicts the layer-wise performance for the three model scales across our text-based classification tasks. We observe:

\begin{itemize}[itemsep=-2pt,topsep=1.5pt]
    \item \textbf{Layer Influence:} Middle-layer activations typically produce slightly higher F1 scores than early- or late-layer features, indicating that mid-level representations strike a useful balance between semantic and syntactic information for classification tasks.
    \item \textbf{Model Scale:} Larger models (9B, 9B-IT) achieve consistently higher mean performance (above 0.85 F1) compared to the 2B model. This aligns with larger hidden dimension in these models having richer representations.
    \item \textbf{SAE Outperforms Baselines:} SAE based features often exceed the performance of the TF-IDF baseline (dotted black line) and final-hidden-state probe (red dashed line)
\end{itemize}

\subsection{Pooling Strategies and Binarization}

We next examine pooling and binarization strategies. Token level max activation pooling methods included no max pooling (top-0), top-20, and top-50 features per token. Binarization is applied after token aggregation.  Figure~\ref{fig:pooling}(b) compares two feature selection strategies: (1) no max pooling with summation of \emph{all} SAE features, and (2) selecting the top-$N$ token level activations (here, 20 and 50), with and without binarization. LR classifiers are trained on the resulting features with L2 regularization.
\begin{itemize}[itemsep=-1.7pt,topsep=1.5pt]
    \item \textbf{Binarization:}  Binarized and no max pooling of SAE features outperform both hidden-state probes and bag-of-words (dotted lines in Figure~\ref{fig:pooling}(b)).  This indicates the effectiveness of SAE features, particularly when combined with binarization, for capturing relevant information.
    \item \textbf{Token level top-$N$ Selection:} Can outperform the binarized and no max pooling approach in certain settings, especially when $N$ increases, and not binarized. However, the margin is typically small, and top-$N$ selection demands additional computation to identify discriminative features.
\end{itemize}

These observations motivate our decision to adopt binarized and no max pooling as a default due to theoretical reduced computational overhead whilst maintaining performance, while acknowledging that token-level top-$N$ might excel for certain tasks.
\vspace{-3pt}

\paragraph{Interpretability and Layer-Wise Insights:}
We find that \emph{middle-layer} SAE features often produce the highest accuracy across tasks. This trend echoes prior work suggesting that intermediate layers encode richer, more compositional representations than either early or late layers. Crucially, we find that binarizing the full set of SAE features offers a robust one-size-fits-all approach, whereas selecting a top-$N$ subset can yield slightly higher performance but requires additional computational steps. From an interpretability perspective, the binarization strategy also grants a straightforward notion of “feature activation”: whether or not a feature dimension was triggered above zero. Such a thresholding approach can facilitate more useful and usable feature-level analyses and potential explanations for model decisions.

\subsection{Cross-Modal Transfer of Text-Trained SAE Features}

Finally, we conduct a preliminary investigation into the cross-modal applicability of SAE features trained on text. Specifically, we tested whether features useful for text classification could also be beneficial in a vision-language setting.

\paragraph{Experimental Setup:}
Instead of using text-based Gemma models directly, we use a Gemma-based LLaVa model (\emph{PaliGemma 2}) \cite{liu2023improvedllava}, which processes both image and text inputs. Activations from image-text pairs were fed into a Gemma-based SAE of equivalent size to assess whether a text-trained SAE could extract meaningful features from multimodal representations. We then classified images from CIFAR-100 \cite{krizhevsky2009learning}, Indian food \cite{rajistics_indian_food_images}, and Oxford Flowers \cite{Nilsback08} using SAE-derived features.

\paragraph{SAE Features Transfer Modalities Effectively:}
The results of these cross-modal experiments are detailed in Appendix \ref{app:multimodal}. We found that the binarization and no max pooling strategy, effective for text-only tasks, remained effective with SAE features derived from \emph{PaliGemma 2} processing partial textual inputs in a vision-language environment. While these initial findings are promising, a more comprehensive study tailored for multimodal analysis is needed to fully explore the benefits and limitations of transferring text-trained SAE features to vision-language tasks.

\section{Multilingual Classification and Transferability}
\label{sec:multilingual}
This section evaluates the cross-lingual robustness of SAE features. We investigate whether features extracted from multilingual datasets are consistent with those found in monolingual contexts and explore the correlation between SAE feature transferability and cross-lingual prediction performance.  We conduct three primary experiments: (1) comparing native and cross-lingual transfer, (2) evaluating different feature selection methods, and (3) assessing the impact of training data sampling.

\paragraph{Dataset:}
We use the multilingual toxicity detection dataset \cite{dementieva2024overview}, which contains text in five languages labeled with a binary toxicity label: English (EN), Chinese (ZH), French (FR), Spanish (ES), and Russian (RU).

\subsection{Native vs. Cross-Lingual Transfer}

\begin{figure}[t]
    \centering
    \includegraphics[width=\linewidth]{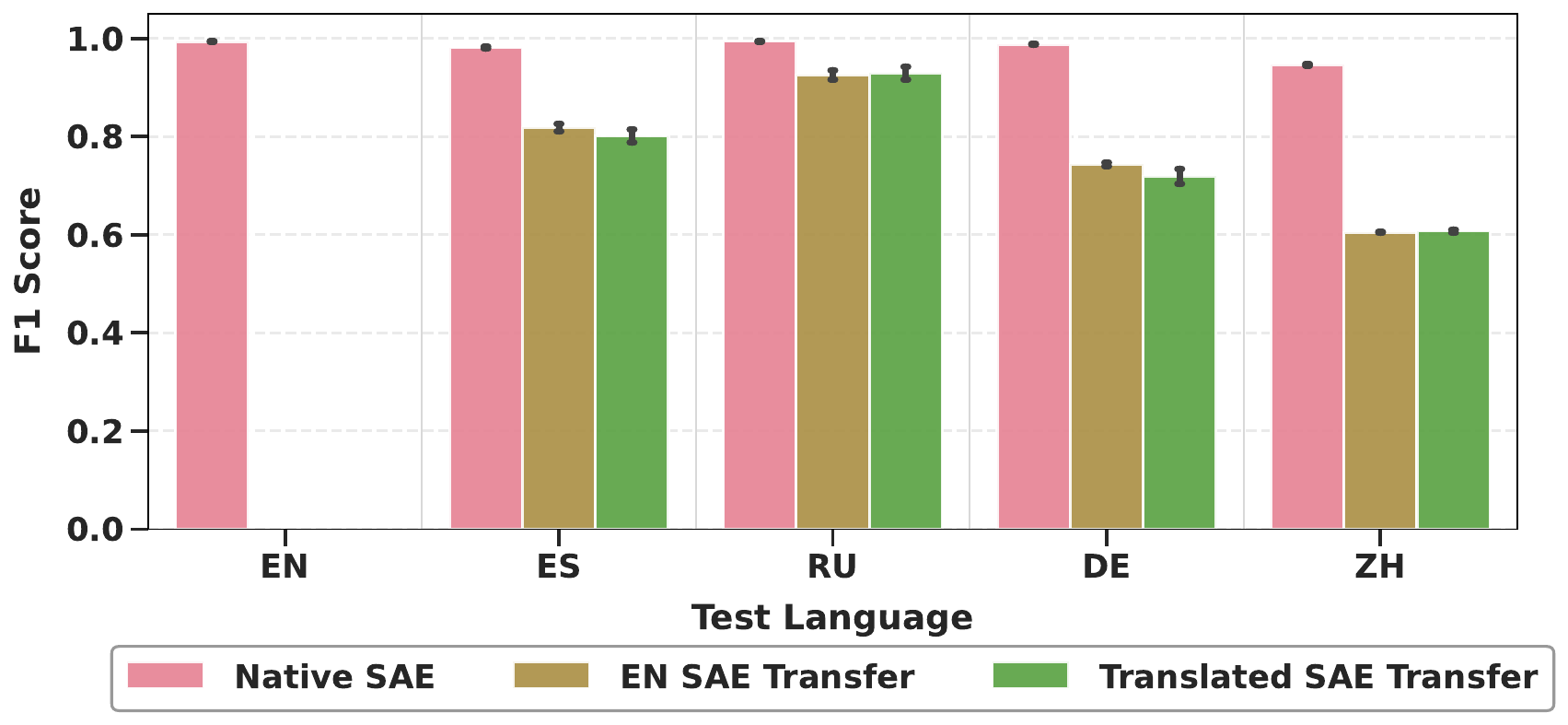}
    \vspace{-0.2cm} 
    \caption{Multilingual toxicity detection results (middle-layer features): \textbf{Native SAE Training} (pink) consistently achieves the best F1 scores. Transferring from English (gold) or using translated inputs (green) leads to moderate performance declines. 9B-IT models show a similar trend, with slightly improved cross-lingual generalization in some language pairs.}
    \label{fig:multilingual}
\end{figure}

We first investigate the performance of SAE features when training and testing on the same language (native) versus training on one language and testing on another (cross-lingual).

\paragraph{Experimental Setup:}
Following the best configurations from previous Section, we extract SAE features from \texttt{gemma-2-9b} and \texttt{9b-it} (widths of 16K or 131K). We train linear classifiers on one language's data and test on the same or a different language. We also compare against a simpler SAE feature selection approach, the \emph{top-n mean-difference} baseline (Mean-Diff) \cite{sae_probing}, to determine if the entire feature set is necessary.

\paragraph{Results and Discussion:}
Figure~\ref{fig:multilingual} presents the F1 scores. Pink bars show \emph{native SAE training}, gold bars show English-trained models tested on other languages, and green bars show English-translated models tested on translated inputs:

\begin{itemize}[itemsep=-1.7pt,topsep=1.5pt]
    \item \textbf{Native Training Superiority:} Native training consistently yields the highest F1 scores (e.g., EN $\to$ EN can reach over 0.99 F1).
    \item \textbf{English Transfer Effectiveness:} Transferring SAE features trained on English (gold bars) achieves reasonable performance on ES, RU, and DE, but with a 15-20\% F1 score decrease compared to native training. This indicates some cross-lingual features generalization internally inside of the models.
    \item \textbf{Direct Transfer Outperforms Translation:} Translating foreign language inputs into English before classification \textbf{does not} outperform direct training on the original language data. Native language signals can be effectively transferred into a shared SAE feature space, proving valuable even without explicit translation.
\end{itemize}

These results suggest that SAE-based representations have cross-lingual potential, but \emph{native} training remains superior. Instruction tuning (\texttt{9B-IT}) yields modest gains, implying distributional shifts from instruction tuning may improve adaptability. Notably, an English-trained SAE performs well in related languages, even better than translations.

\subsection{Feature Selection Methods: Full SAE vs. Hidden States vs. Mean-Diff}

\begin{figure}[t]
    \centering
    \includegraphics[width=\linewidth]{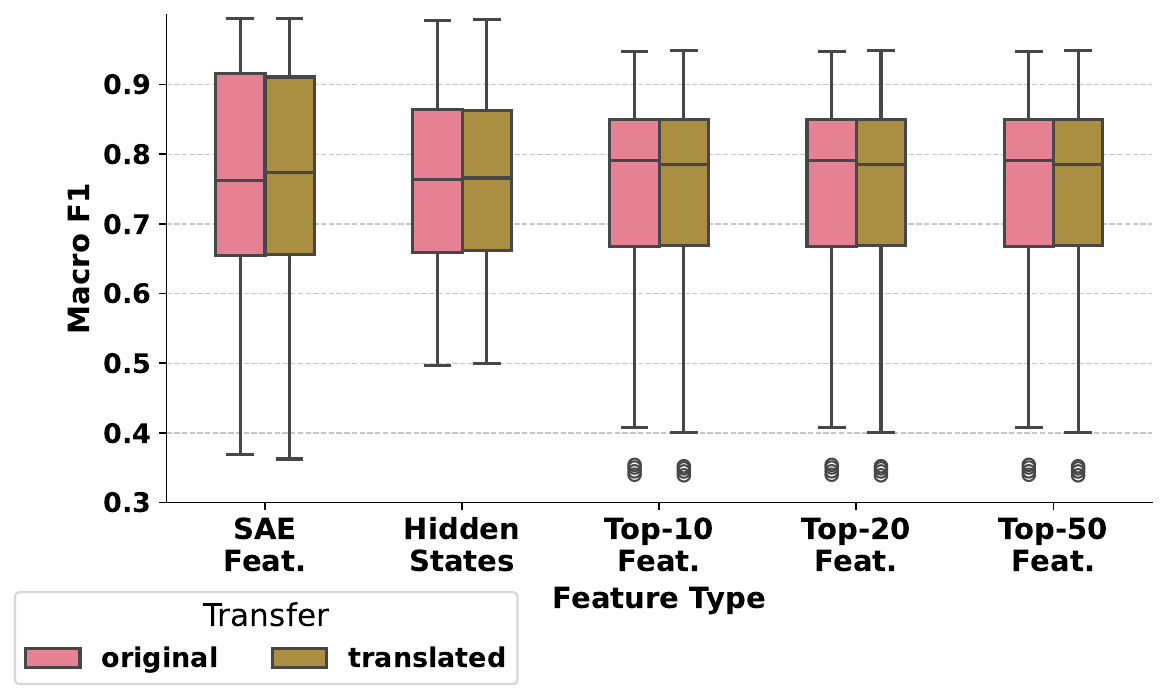}
    \caption{Comparison of average F1 scores by different feature selection methods on the Multilingual Classification and Transfer task. The boxes represent the mean $\pm$ standard deviation, and the whiskers indicate the interquartile range (IQR).}
    \label{fig:boxplot}
\end{figure}


\paragraph{Experimental Setup:}
We compare feature selection methods on \texttt{gemma-2-9b} and \texttt{9b-it}, analyzing performance across different layers using: all SAE features (with binarization), last token hidden-state probing (baseline), and the top-$N$ mean-difference (Mean-Diff) approach.

\paragraph{Results and Discussion:}
Figure~\ref{fig:boxplot} shows the average F1 scores across layers.\footnote{Large variance of the box plot here are caused by transfer across 5 languages and 3 layer settings within 2 models.} \textbf{SAE features achieve the highest macro F1 scores} but exhibit \textbf{greater variance}, particularly due to DE → ZH transfer. Despite this, they remain the \textbf{most preferable choice} due to their superior peak performance. \textbf{Hidden-state probing} performs competitively with \textbf{lower variance} but does not reach the highest scores, making it a more stable alternative. Meanwhile, \textbf{Mean-Diff top-$N$ selection} (Top-10, Top-20, Top-50) consistently lags behind SAE features and hidden states, offering \textbf{similar variance but lower effectiveness}, reinforcing the benefit of using the full SAE feature set.\footnote{These different methods also utilize different important features to do classification which results in performance differences as shown in Appendix \ref{app:cross_lingual_features}.}

However, when considering average rather than peak performance, Mean-Diff top-N selection actually outperforms SAE features, providing a higher mean F1 score and lower variance. This suggests it may be preferable in scenarios where stability across tasks is prioritized over peak performance. We then examine the robustness of SAE feature extraction with varying amounts of training data.

\paragraph{Experimental Setup:}
We assess performance across training set sampling rates (0.1–1.0), comparing native language training and English transfer. For each, we evaluate SAE binarized features, hidden states, and Mean-Diff top-$N$ selection.

\paragraph{Results and Discussion:}
Figure~\ref{fig:sampling-main} shows the performance across sampling rates. Key findings:

\begin{itemize}\itemsep=-1.7pt
    \item \textbf{Native Outperforms Transfer:} Native language training consistently outperforms English transfer across \textbf{all sampling rates}.
    \item \textbf{SAE Features are Superior:} Our full binarized SAE features achieve superior F1 scores (0.85-0.90) compared to both hidden states (0.80-0.85) and top-$N$ selection (0.75-0.80).
    \item \textbf{Stable Performance Gap:} The performance difference between native and transfer settings remains relatively stable even with limited data. This shows that our feature extraction method is robust even when data is scarce.
\end{itemize}

\paragraph{Clarifying differences from \cite{sae_probing}}
The use of L1 sparsity methods to perform feature selection, mean-difference approach of \cite{sae_probing}, demonstrates strong performance and that a small number of features can contain most of the task-relevant information. However, for the specific task of our multilingual toxicity detection, the aggregated binarisation method from all features appears to preserve a stronger signal and greater transferability across languages in native and translated settings. This is in contrast to Kantamneni et al. (2025), and therefore, future work is needed to clarify the task sensitivity of the divergent findings. Major differences on feature selection methods may also drive differences and future work will focus on understanding the impact of different methods on varied interpretability approaches.

\section{Behavioral (Action) Prediction}
\label{sec:action}
This section examines whether smaller models can predict the output correctness ("action") of larger, instruction-tuned models in knowledge-intensive QA tasks. This relates to \emph{scalable oversight}, where a smaller, interpretable model monitors a more capable system. We focus on predicting the \texttt{9B-IT} model's behavior using features from smaller models and assess the impact of context fidelity.

\paragraph{Goal and Motivation:}
We aim to determine whether smaller and/or base models (\texttt{Gemma 2-2B, 9B}) can predict their own behavior or that of a larger and/or fine-tuned model (\texttt{9B-IT}) on knowledge-based QA tasks, based on correct or incorrect factual information. This aligns with a \emph{scalable oversight} scenario, where a smaller model monitors a more capable system when they share the same corpus and architecture.
\vspace{-0.2em}

\paragraph{Datasets:}
We use the entity-based knowledge conflicts in question answering dataset \cite{longpre2022entitybasedknowledgeconflictsquestion}, which provides binary correctness labels for model responses. Open-ended generation is performed with \emph{vllm} \cite{kwon2023efficient}, and answers are scored using \emph{inspect ai} \cite{UKGovernmentBEIS_inspect_ai} with GPT-4o-mini as the grader.
\vspace{-0.2em}

\begin{figure}[t]
    \centering
    \includegraphics[width=\linewidth]{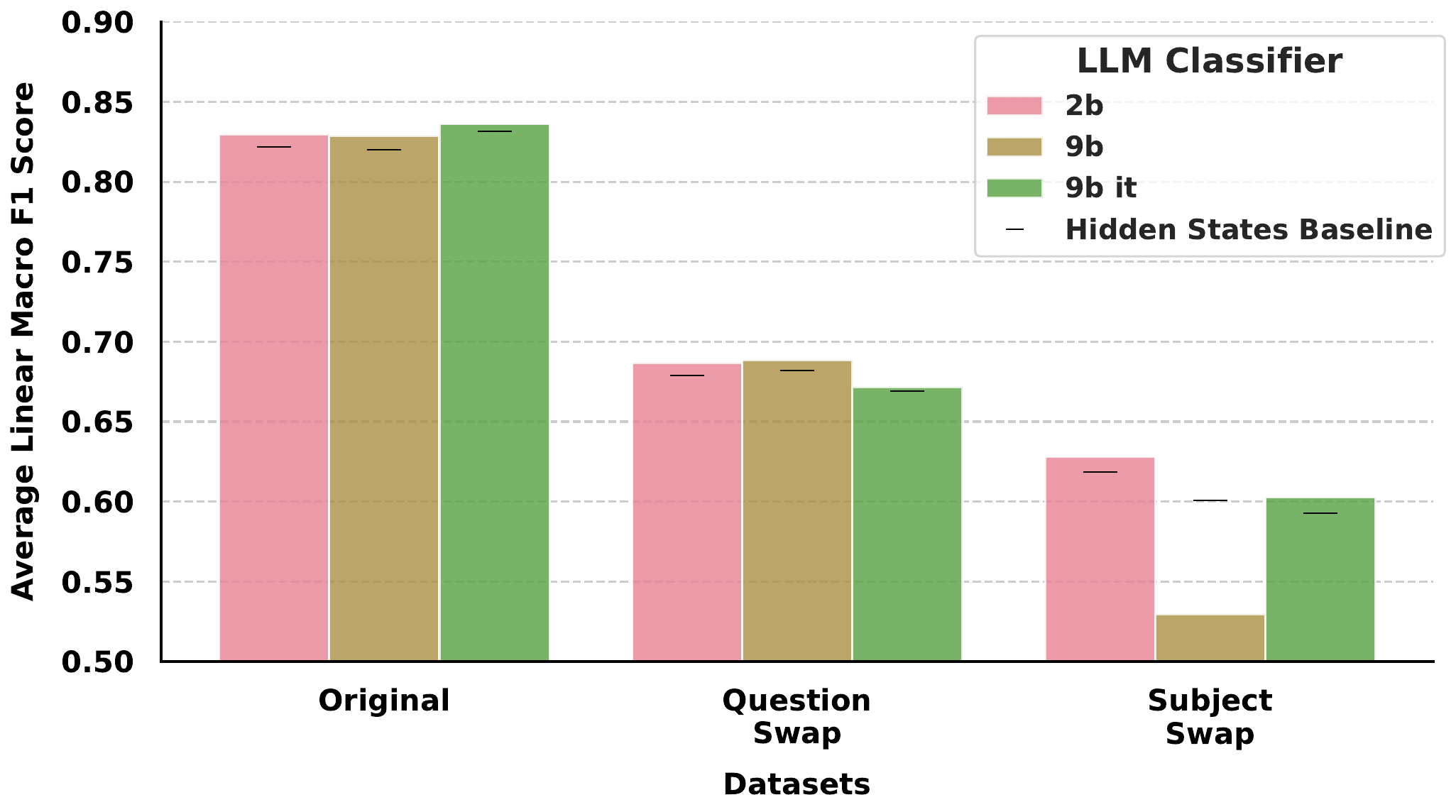}
    \caption{Action prediction performance for \texttt{9B-IT} across different context manipulations (Original, Question Swap, Subject Swap). Each bar represents a different LLM extracted features trained into classifiers (\texttt{2B, 9B, 9B-IT}) using SAE features; the black horizontal lines indicate the hidden-states baseline. High predictive power is observed with the correct context, dropping significantly with context manipulations. 2B-based features are competitive in predicting 9B-IT’s behaviors.}
    \label{fig:action}
\end{figure}
\vspace{-0.2em}

\paragraph{Experimental Design and Results:}
We focus on predicting \texttt{9B-IT}'s output correctness. For a given model $M$ (\texttt{2B, 9B, 9B-IT}): \textbf{1)} We generate open-ended answers to prompts using the model. \textbf{2)} We use \texttt{GPT-4o-mini-0718} to label each answer as correct or incorrect. \textbf{3)} We extract pretrained SAE activations from the input question, with and without provided contexts. \textbf{4)} We train a logistic regression model to predict the binary correctness label from these extracted features.

We also perform cross-model prediction (e.g., \texttt{2B} predicting \texttt{9B}'s correctness), similar to \cite{binder2024lookinginwardlanguagemodels}. We fix the SAE width to 16K and compare the quality of predictions using full SAE binarized approach to those using the Top-$N$ mean difference feature method, and analyze auto-interpretable descriptions of features to understand if similar explanations are shared in the top features across models. Figure~\ref{fig:action} summarizes the macro F1 scores across several conditions from the \emph{NQ-Swap} and \emph{inspect\_evals} datasets: Original context, Question Swap, and Subject Swap. Key findings:

\begin{itemize}[itemsep=-1.7pt,topsep=1.5pt]
    \item \textbf{Context Fidelity is Crucial:} Providing the correct context ("Original" setting) yields the highest F1 scores (above 80\%). Removing or swapping the context causes a significant drop (20\%), underscoring the importance of reliable response prediction across contexts.
    \item \textbf{Inter-Model Prediction is Effective:} Surprisingly, 2B-based SAE features can predict \texttt{9B-IT}'s correctness nearly as well as, and sometimes \emph{better than}, \texttt{9B-IT}'s \emph{own} features. This is a key result for scalable oversight.
    \item \textbf{SAE Features Outperform Hidden States:} Hidden-state baselines (black lines) generally perform worse than the binarized SAE feature sets, reinforcing the utility of the SAE-based approach for this "behavior prediction" task.
\end{itemize}

\paragraph{Implications for Scalable Oversight:}
These findings highlight the promise of using smaller SAEs to interpret or predict the actions of more powerful LMs. Although context consistency is critical, the ability to forecast a larger model's decisions has significant implications for AI safety and governance, especially in risk-sensitive domains.
In summary, our results demonstrate that:
\begin{enumerate}[itemsep=-1.7pt,topsep=1.5pt]
    \item SAE-based features consistently outperform hidden-state and TF-IDF baselines across classification tasks, especially when using summation + binarization.
    \item For multilingual toxicity detection, native training outperforms cross-lingual transfer, though instruction-tuned models (e.g., \texttt{9B-IT}) may exhibit modestly better transfer as you can see in Appendix \ref{app:action_paired_plot} and \ref{app:action_autointerp}.
    \item Smaller LMs can leverage SAE features to accurately predict the behavior of larger instruction-tuned models, suggesting a scalable mechanism for oversight and auditing.
\end{enumerate}

\section{Conclusion}
\label{sec:conclusion}
We present a comprehensive study of SAE features across multiple model scales, tasks, languages, and modalities, highlighting both their practical strengths and interpretive advantages. Specifically, summation-then-binarization of SAE features surpassed hidden-state probes and bag-of-words baselines in most tasks, while demonstrating cross-lingual transferability. Moreover, we showed that smaller LLMs equipped with SAE features can effectively predict the actions of larger models, pointing to a potential mechanism for \emph{scalable oversight} and auditing. Taken together, these results reinforce the idea that learning (or adopting) a sparse, disentangled representation of internal activations can yield significant performance benefits and support interpretability objectives.

We hope this work will serve as a foundation for future studies that exploit SAEs in broader multimodal, diverse languages, and complex real-world workflows where trust and accountability are paramount. By marrying strong classification performance with clearer feature-level insights, SAE-based methods represent a promising path toward safer and more transparent LLM applications.

\section{Limitations}

While our study demonstrates the effectiveness of SAE features for classification and transferability, several limitations remain.

\paragraph{Dependence on Gemma 2 Pretrained-SAEs}  
Our primary analysis is restricted to SAEs trained with Jump ReLU activation on Gemma 2 models as they were the only open-source models available that provided SAE’s across varying layers, widths, and model sizes. This could potentially limit generalizability to other model architectures and training paradigms. Future work should explore diverse SAE training strategies and model sources.

\paragraph{Limited Multimodal and Cross-Lingual Evaluation}  
Our cross-modal experiments are preliminary, and further research is needed to validate SAE generalization across different modalities and low-resource languages.

\paragraph{Sensitivity to Task and Data Distribution}  
SAE performance varies across datasets, and its robustness under adversarial conditions or domain shifts needs further study.

\paragraph{Interpretability Challenges}  
Despite improved feature transparency, the semantic alignment of SAE features with human-interpretable concepts remains an open question.

\paragraph{Potential Risks}
The toxicity or other safety-related open-sourced data we use contained offensive language, which we have not shown in the manuscript. And the auto-interp features are fully AI generated by neuronpedia.org.

\paragraph{Future Work: Robustness under Domain Shift}
A crucial next step is to investigate how SAE-derived features behave when the input distribution changes. This includes examining covariate, subpopulation, and temporal shifts by training probes on one domain and evaluating on held-out domains (e.g., news→social media; formal→informal), measuring activation drift and the stability of feature–label associations. This evaluation will clarify whether the observed transferability reflects domain-agnostic structure or domain-specific correlations.

Beyond robustness, there is a need to expand the task set beyond safety-oriented binary classification to include multilabel and non-safety tasks and additional multilingual benchmarks.

\section*{Acknowledgments}
The authors acknowledge financial support from the Google PhD Fellowship (SC), the Woods Foundation (DB, SC, JG), the NIH (NIH R01CA294033 (SC, JG, DB), NIH U54CA274516-01A1 (SC, DB) and the American Cancer Society and American Society for Radiation Oncology, ASTRO-CSDG-24-1244514-01-CTPS Grant DOI: ACS.ASTRO-CSDG-24-1244514-01-CTPS.pc.gr.222210 (DB).

The authors extend their gratitude to John Osborne from UAB for his support and to Zidi Xiong from Harvard for proofreading the preprint. Author SC also appreciates the advice on this project from Fred Zhang and Asma Ghandeharioun from Google through the mentorships program.

\bibliography{references}

@misc{hinck2024llavagemma,
      title={LLaVA-Gemma: Accelerating Multimodal Foundation Models with a Compact Language Model}, 
      author={Musashi Hinck and Matthew L. Olson and David Cobbley and Shao-Yen Tseng and Vasudev Lal},
      year={2024},
      eprint={2404.01331},
      url={https://arxiv.org/abs/2404.01331},
      archivePrefix={arXiv},
      primaryClass={cs.CL}
}

@misc{muennighoff2023mtebmassivetextembedding,
      title={MTEB: Massive Text Embedding Benchmark}, 
      author={Niklas Muennighoff and Nouamane Tazi and Loïc Magne and Nils Reimers},
      year={2023},
      eprint={2210.07316},
      archivePrefix={arXiv},
      primaryClass={cs.CL},
      url={https://arxiv.org/abs/2210.07316}, 
}

@misc{karvonen2025saebenchcomprehensivebenchmarksparse,
      title={SAEBench: A Comprehensive Benchmark for Sparse Autoencoders in Language Model Interpretability}, 
      author={Adam Karvonen and Can Rager and Johnny Lin and Curt Tigges and Joseph Bloom and David Chanin and Yeu-Tong Lau and Eoin Farrell and Callum McDougall and Kola Ayonrinde and Demian Till and Matthew Wearden and Arthur Conmy and Samuel Marks and Neel Nanda},
      year={2025},
      eprint={2503.09532},
      archivePrefix={arXiv},
      primaryClass={cs.LG},
      url={https://arxiv.org/abs/2503.09532}, 
}

@inproceedings{shen2018baseline,
  title={Baseline Needs More Love: On Simple Word-Embedding-Based Models and Associated Pooling Mechanisms},
  author={Shen, Dinghan and Wang, Guoyin and Wang, Wenlin and Min, Martin Renqiang and Su, Qinliang and Zhang, Yizhe and Li, Chunyuan and Henao, Ricardo and Carin, Lawrence},
  booktitle={Proceedings of the 56th Annual Meeting of the Association for Computational Linguistics (Volume 1: Long Papers)},
  pages={440--450},
  year={2018}
}

@article{brinkmann2025large,
  title={Large Language Models Share Representations of Latent Grammatical Concepts Across Typologically Diverse Languages},
  author={Brinkmann, Jannik and Wendler, Chris and Bartelt, Christian and Mueller, Aaron},
  journal={arXiv preprint arXiv:2501.06346},
  year={2025}
}

@misc{agarap2019deeplearningusingrectified,
      title={Deep Learning using Rectified Linear Units (ReLU)}, 
      author={Abien Fred Agarap},
      year={2019},
      eprint={1803.08375},
      archivePrefix={arXiv},
      primaryClass={cs.NE},
      url={https://arxiv.org/abs/1803.08375}, 
}

@article{sparck_jones_1972,
  title = {A Statistical Interpretation of Term Specificity and Its Application in Retrieval},
  author = {Spärck Jones, Karen},
  journal = {Journal of Documentation},
  volume = {28},
  number = {1},
  pages = {11--21},
  year = {1972},
  doi = {10.1108/eb026526},
  url = {https://www.emerald.com/insight/content/doi/10.1108/eb026526/full/html}
}

@misc{liu2023improvedllava,
          author={Liu, Haotian and Li, Chunyuan and Li, Yuheng and Lee, Yong Jae},
          title={Improved Baselines with Visual Instruction Tuning}, 
          publisher={arXiv:2310.03744},
          year={2023},
  }

@inproceedings{Nilsback08,
  author    = {Maria-Elena Nilsback and Andrew Zisserman},
  title     = {Automated Flower Classification over a Large Number of Classes},
  booktitle = {Indian Conference on Computer Vision, Graphics and Image Processing},
  month     = {Dec},
  year      = {2008},
}

@misc{bloom2024saetrainingcodebase,
   title = {SAELens},
   author = {Joseph Bloom, Curt Tigges and David Chanin},
   year = {2024},
   howpublished = {\url{https://github.com/jbloomAus/SAELens}},
}

@misc{nanda2022transformerlens,
    title = {TransformerLens},
    author = {Neel Nanda and Joseph Bloom},
    year = {2022},
    howpublished = {\url{https://github.com/TransformerLensOrg/TransformerLens}},
}

@inproceedings{Casanueva2020,
    author      = {I{\~{n}}igo Casanueva and Tadas Temcinas and Daniela Gerz and Matthew Henderson and Ivan Vulic},
    title       = {Efficient Intent Detection with Dual Sentence Encoders},
    year        = {2020},
    month       = {mar},
    note        = {Data available at https://github.com/PolyAI-LDN/task-specific-datasets},
    url         = {https://arxiv.org/abs/2003.04807},
    booktitle   = {Proceedings of the 2nd Workshop on NLP for ConvAI - ACL 2020}
}

@misc{longpre2022entitybasedknowledgeconflictsquestion,
      title={Entity-Based Knowledge Conflicts in Question Answering}, 
      author={Shayne Longpre and Kartik Perisetla and Anthony Chen and Nikhil Ramesh and Chris DuBois and Sameer Singh},
      year={2022},
      eprint={2109.05052},
      archivePrefix={arXiv},
      primaryClass={cs.CL},
      url={https://arxiv.org/abs/2109.05052}, 
}

@inproceedings{dementieva2024overview,
  title={Overview of the Multilingual Text Detoxification Task at PAN 2024},
  author={Dementieva, Daryna and Moskovskiy, Daniil and Babakov, Nikolay and Ayele, Abinew Ali and Rizwan, Naquee and Schneider, Frolian and Wang, Xintog and Yimam, Seid Muhie and Ustalov, Dmitry and Stakovskii, Elisei and Smirnova, Alisa and Elnagar, Ashraf and Mukherjee, Animesh and Panchenko, Alexander},
  booktitle={Working Notes of CLEF 2024 - Conference and Labs of the Evaluation Forum},
  editor={Guglielmo Faggioli and Nicola Ferro and Petra Galu{\v{s}}{\v{c}}{\'a}kov{\'a} and Alba Garc{\'i}a Seco de Herrera},
  year={2024},
  organization={CEUR-WS.org}
}

@misc{UKGovernmentBEIS_inspect_ai,
  author = {AI Safety Institute, UK},
  title = {Inspect {AI:} {Framework} for {Large} {Language} {Model}
    {Evaluations}},
  date = {2024-05},
  url = {https://github.com/UKGovernmentBEIS/inspect_ai},
  langid = {en}
}

@misc{gemmateam2024gemma2improvingopen,
      title={Gemma 2: Improving Open Language Models at a Practical Size}, 
      author={Gemma Team and Morgane Riviere and Shreya Pathak and Pier Giuseppe Sessa and Cassidy Hardin and Surya Bhupatiraju and Léonard Hussenot and Thomas Mesnard and Bobak Shahriari and Alexandre Ramé and Johan Ferret and Peter Liu and Pouya Tafti and Abe Friesen and Michelle Casbon and Sabela Ramos and Ravin Kumar and Charline Le Lan and Sammy Jerome and Anton Tsitsulin and Nino Vieillard and Piotr Stanczyk and Sertan Girgin and Nikola Momchev and Matt Hoffman and Shantanu Thakoor and Jean-Bastien Grill and Behnam Neyshabur and Olivier Bachem and Alanna Walton and Aliaksei Severyn and Alicia Parrish and Aliya Ahmad and Allen Hutchison and Alvin Abdagic and Amanda Carl and Amy Shen and Andy Brock and Andy Coenen and Anthony Laforge and Antonia Paterson and Ben Bastian and Bilal Piot and Bo Wu and Brandon Royal and Charlie Chen and Chintu Kumar and Chris Perry and Chris Welty and Christopher A. Choquette-Choo and Danila Sinopalnikov and David Weinberger and Dimple Vijaykumar and Dominika Rogozińska and Dustin Herbison and Elisa Bandy and Emma Wang and Eric Noland and Erica Moreira and Evan Senter and Evgenii Eltyshev and Francesco Visin and Gabriel Rasskin and Gary Wei and Glenn Cameron and Gus Martins and Hadi Hashemi and Hanna Klimczak-Plucińska and Harleen Batra and Harsh Dhand and Ivan Nardini and Jacinda Mein and Jack Zhou and James Svensson and Jeff Stanway and Jetha Chan and Jin Peng Zhou and Joana Carrasqueira and Joana Iljazi and Jocelyn Becker and Joe Fernandez and Joost van Amersfoort and Josh Gordon and Josh Lipschultz and Josh Newlan and Ju-yeong Ji and Kareem Mohamed and Kartikeya Badola and Kat Black and Katie Millican and Keelin McDonell and Kelvin Nguyen and Kiranbir Sodhia and Kish Greene and Lars Lowe Sjoesund and Lauren Usui and Laurent Sifre and Lena Heuermann and Leticia Lago and Lilly McNealus and Livio Baldini Soares and Logan Kilpatrick and Lucas Dixon and Luciano Martins and Machel Reid and Manvinder Singh and Mark Iverson and Martin Görner and Mat Velloso and Mateo Wirth and Matt Davidow and Matt Miller and Matthew Rahtz and Matthew Watson and Meg Risdal and Mehran Kazemi and Michael Moynihan and Ming Zhang and Minsuk Kahng and Minwoo Park and Mofi Rahman and Mohit Khatwani and Natalie Dao and Nenshad Bardoliwalla and Nesh Devanathan and Neta Dumai and Nilay Chauhan and Oscar Wahltinez and Pankil Botarda and Parker Barnes and Paul Barham and Paul Michel and Pengchong Jin and Petko Georgiev and Phil Culliton and Pradeep Kuppala and Ramona Comanescu and Ramona Merhej and Reena Jana and Reza Ardeshir Rokni and Rishabh Agarwal and Ryan Mullins and Samaneh Saadat and Sara Mc Carthy and Sarah Cogan and Sarah Perrin and Sébastien M. R. Arnold and Sebastian Krause and Shengyang Dai and Shruti Garg and Shruti Sheth and Sue Ronstrom and Susan Chan and Timothy Jordan and Ting Yu and Tom Eccles and Tom Hennigan and Tomas Kocisky and Tulsee Doshi and Vihan Jain and Vikas Yadav and Vilobh Meshram and Vishal Dharmadhikari and Warren Barkley and Wei Wei and Wenming Ye and Woohyun Han and Woosuk Kwon and Xiang Xu and Zhe Shen and Zhitao Gong and Zichuan Wei and Victor Cotruta and Phoebe Kirk and Anand Rao and Minh Giang and Ludovic Peran and Tris Warkentin and Eli Collins and Joelle Barral and Zoubin Ghahramani and Raia Hadsell and D. Sculley and Jeanine Banks and Anca Dragan and Slav Petrov and Oriol Vinyals and Jeff Dean and Demis Hassabis and Koray Kavukcuoglu and Clement Farabet and Elena Buchatskaya and Sebastian Borgeaud and Noah Fiedel and Armand Joulin and Kathleen Kenealy and Robert Dadashi and Alek Andreev},
      year={2024},
      eprint={2408.00118},
      archivePrefix={arXiv},
      primaryClass={cs.CL},
      url={https://arxiv.org/abs/2408.00118}, 
}

@inproceedings{kwon2023efficient,
  title={Efficient Memory Management for Large Language Model Serving with PagedAttention},
  author={Woosuk Kwon and Zhuohan Li and Siyuan Zhuang and Ying Sheng and Lianmin Zheng and Cody Hao Yu and Joseph E. Gonzalez and Hao Zhang and Ion Stoica},
  booktitle={Proceedings of the ACM SIGOPS 29th Symposium on Operating Systems Principles},
  year={2023}
}

@misc{binder2024lookinginwardlanguagemodels,
      title={Looking Inward: Language Models Can Learn About Themselves by Introspection}, 
      author={Felix J Binder and James Chua and Tomek Korbak and Henry Sleight and John Hughes and Robert Long and Ethan Perez and Miles Turpin and Owain Evans},
      year={2024},
      eprint={2410.13787},
      archivePrefix={arXiv},
      primaryClass={cs.CL},
      url={https://arxiv.org/abs/2410.13787}, 
}

@misc{anthropic_election_questions,
  author = {Anthropic},
  title = {Election Questions Dataset},
  year = {2023},
  url = {https://huggingface.co/datasets/Anthropic/election_questions}
}

@article{arditi2024refusal,
  title={Refusal in Language Models Is Mediated by a Single Direction},
  author={Arditi, Andy and Obeso, Oscar and Syed, Aaquib and Paleka, Daniel and Panickssery, Nina and Gurnee, Wes and Nanda, Neel},
  journal={arXiv preprint arXiv:2406.11717},
  year={2024},
  url={https://arxiv.org/abs/2406.11717}
}

@misc{jackhhao_jailbreak_classification,
  author = {Hao, Jack},
  title = {Jailbreak Classification Dataset},
  year = {2023},
  url = {https://huggingface.co/datasets/jackhhao/jailbreak-classification}
}

@inproceedings{fitzgerald2023massive,
  title = {{MASSIVE}: A 1M-Example Multilingual Natural Language Understanding Dataset with 51 Typologically-Diverse Languages},
  author = {FitzGerald, Jack and Hench, Christopher and Peris, Charith and Mackie, Scott and Rottmann, Kay and Sanchez, Ana and Nash, Aaron and Urbach, Liam and Kakarala, Vishesh and Singh, Richa and Ranganath, Swetha and Crist, Laurie and Britan, Misha and Leeuwis, Wouter and Tur, Gokhan and Natarajan, Prem},
  booktitle = {Proceedings of the 61st Annual Meeting of the Association for Computational Linguistics (Volume 1: Long Papers)},
  year = {2023},
  url = {https://aclanthology.org/2023.acl-long.235/}
}

@misc{setfit_tweet_eval_stance_abortion,
  author = {SetFit},
  title = {TweetEval Stance Abortion Dataset},
  year = {2023},
  url = {https://huggingface.co/datasets/SetFit/tweet_eval_stance_abortion}
}

@article{krizhevsky2009learning,
  title={Learning multiple layers of features from tiny images},
  author={Krizhevsky, Alex and Hinton, Geoffrey},
  year={2009},
  institution={University of Toronto},
  url={https://www.cs.toronto.edu/~kriz/cifar.html}
}

@misc{nelorth_oxford_flowers,
  author = {Nelorth},
  title = {Oxford Flowers Dataset},
  year = {2023},
  url = {https://huggingface.co/datasets/nelorth/oxford-flowers}
}

@misc{rajistics_indian_food_images,
  author = {Rajistics},
  title = {Indian Food Images Dataset},
  year = {2023},
  url = {https://huggingface.co/datasets/rajistics/indian_food_images}
}

@article{sae_probing,
  title={SAE Probing: What Is It Good For? Absolutely Something!},
  author={Kantamneni, Subhash and Engels, Josh and Rajamanoharan, Senthooran and Nanda, Neel},
  journal={LessWrong},
  year={2024},
  url={https://www.lesswrong.com/posts/NMLq8yoTecAF44KX9}
}

@article{sae_features_llava,
  title={Are SAE Features from the Base Model Still Meaningful to LLaVA?},
  author={Chen, Shan and Gallifant, Jack and Sasse, Kuleen and Bitterman, Danielle},
  journal={LessWrong},
  year={2024},
  url={https://www.lesswrong.com/posts/8JTi7N3nQmjoRRuMD}
}

@misc{bowman2022measuringprogressscalableoversight,
      title={Measuring Progress on Scalable Oversight for Large Language Models}, 
      author={Samuel R. Bowman and Jeeyoon Hyun and Ethan Perez and Edwin Chen and Craig Pettit and Scott Heiner and Kamilė Lukošiūtė and Amanda Askell and Andy Jones and Anna Chen and Anna Goldie and Azalia Mirhoseini and Cameron McKinnon and Christopher Olah and Daniela Amodei and Dario Amodei and Dawn Drain and Dustin Li and Eli Tran-Johnson and Jackson Kernion and Jamie Kerr and Jared Mueller and Jeffrey Ladish and Joshua Landau and Kamal Ndousse and Liane Lovitt and Nelson Elhage and Nicholas Schiefer and Nicholas Joseph and Noemí Mercado and Nova DasSarma and Robin Larson and Sam McCandlish and Sandipan Kundu and Scott Johnston and Shauna Kravec and Sheer El Showk and Stanislav Fort and Timothy Telleen-Lawton and Tom Brown and Tom Henighan and Tristan Hume and Yuntao Bai and Zac Hatfield-Dodds and Ben Mann and Jared Kaplan},
      year={2022},
      eprint={2211.03540},
      archivePrefix={arXiv},
      primaryClass={cs.HC},
      url={https://arxiv.org/abs/2211.03540}, 
}

@misc{abdulaal2024xrayworth15features,
      title={An X-Ray Is Worth 15 Features: Sparse Autoencoders for Interpretable Radiology Report Generation}, 
      author={Ahmed Abdulaal and Hugo Fry and Nina Montaña-Brown and Ayodeji Ijishakin and Jack Gao and Stephanie Hyland and Daniel C. Alexander and Daniel C. Castro},
      year={2024},
      eprint={2410.03334},
      archivePrefix={arXiv},
      primaryClass={cs.CV},
      url={https://arxiv.org/abs/2410.03334}, 
}

@article{bricken2023towards,
  title={Towards Monosemanticity: Decomposing Language Models With Dictionary Learning},
  author={Trenton Bricken and Adly Templeton and Joshua Batson and Brian Chen and Adam Jermyn and Tom Conerly and Nicholas L Turner and Cem Anil and Carson Denison and Amanda Askell and Robert Lasenby and Yifan Wu and Shauna Kravec and Nicholas Schiefer and Tim Maxwell and Nicholas Joseph and Alex Tamkin and Karina Nguyen and Brayden McLean and Josiah E Burke and Tristan Hume and Shan Carter and Tom Henighan and Chris Olah},
  journal={transformer-circuits.pub},
  year={2023},
  volume={monosemantic-features},
  url={https://transformer-circuits.pub/2023/monosemantic-features}
}

@misc{wu2025axbenchsteeringllmssimple,
      title={AxBench: Steering LLMs? Even Simple Baselines Outperform Sparse Autoencoders}, 
      author={Zhengxuan Wu and Aryaman Arora and Atticus Geiger and Zheng Wang and Jing Huang and Dan Jurafsky and Christopher D. Manning and Christopher Potts},
      year={2025},
      eprint={2501.17148},
      archivePrefix={arXiv},
      primaryClass={cs.CL},
      url={https://arxiv.org/abs/2501.17148}, 
}

@misc{anthropic2024features,
  title={Using Dictionary Learning Features as Classifiers},
  author={Trenton Bricken and Jonathan Marcus and Siddharth Mishra-Sharma and Meg Tong and Ethan Perez and Mrinank Sharma and Kelley Rivoire and Thomas Henighan and Adam Jermyn},
  journal={Transformer Circuits},
  year={2024},
  url={https://transformer-circuits.pub/2024/features-as-classifiers/index.html}
}

@article{marks2024sparse,
  title={Sparse feature circuits: Discovering and editing interpretable causal graphs in language models},
  author={Marks, Samuel and Rager, Can and Michaud, Eric J and Belinkov, Yonatan and Bau, David and Mueller, Aaron},
  journal={arXiv preprint arXiv:2403.19647},
  year={2024}
}

@misc{gao2024scalingevaluatingsparseautoencoders,
      title={Scaling and evaluating sparse autoencoders}, 
      author={Leo Gao and Tom Dupré la Tour and Henk Tillman and Gabriel Goh and Rajan Troll and Alec Radford and Ilya Sutskever and Jan Leike and Jeffrey Wu},
      year={2024},
      eprint={2406.04093},
      archivePrefix={arXiv},
      primaryClass={cs.LG},
      url={https://arxiv.org/abs/2406.04093}, 
}

@misc{elhage2022toy,
      title={Toy Models of Superposition}, 
      author={Nelson Elhage and Tristan Hume and Catherine Olsson and Nicholas Schiefer and Tom Henighan and Shauna Kravec and Zac Hatfield-Dodds and Robert Lasenby and Dawn Drain and Carol Chen and Roger Grosse and Sam McCandlish and Jared Kaplan and Dario Amodei and Martin Wattenberg and Christopher Olah},
      year={2022},
      eprint={2209.10652},
      archivePrefix={arXiv},
      primaryClass={cs.LG},
      url={https://arxiv.org/abs/2209.10652}, 
}

@article{cunningham2023sparse,
  title={Sparse autoencoders find highly interpretable features in language models},
  author={Cunningham, Hoagy and Ewart, Aidan and Riggs, Logan and Huben, Robert and Sharkey, Lee},
  journal={arXiv preprint arXiv:2309.08600},
  year={2023}
}

@article{templeton2024scaling,
  title={Scaling Monosemanticity: Extracting Interpretable Features from Claude 3 Sonnet},
  author={Templeton, A. and Conerly, T. and Marcus, J. and Lindsey, J. and Bricken, T. and Chen, B. and Pearce, A. and Citro, C. and Ameisen, E. and Jones, A. and Cunningham, H. and Turner, N. L. and McDougall, C. and MacDiarmid, M. and Freeman, C. D. and Sumers, T. R. and Rees, E. and Batson, J. and Jermyn, A. and Carter, S. and Olah, C. and Henighan, T.},
  journal={Transformer Circuits Thread},
  year={2024},
  url={https://www.anthropic.com/transformer-circuits}
}

@article{cammarata2021curve,
  title={Curve circuits},
  author={Cammarata, Nick and Goh, Gabriel and Carter, Shan and Voss, Chelsea and Schubert, Ludwig and Olah, Chris},
  journal={Distill},
  volume={6},
  number={1},
  pages={e00024--006},
  year={2021}
}

@article{wang2022interpretability,
  title={Interpretability in the wild: a circuit for indirect object identification in gpt-2 small},
  author={Wang, Kevin and Variengien, Alexandre and Conmy, Arthur and Shlegeris, Buck and Steinhardt, Jacob},
  journal={arXiv preprint arXiv:2211.00593},
  year={2022}
}

@misc{ngo2022alignment,
      title={The Alignment Problem from a Deep Learning Perspective}, 
      author={Richard Ngo and Lawrence Chan and Sören Mindermann},
      year={2025},
      eprint={2209.00626},
      archivePrefix={arXiv},
      primaryClass={cs.AI},
      url={https://arxiv.org/abs/2209.00626}, 
}

@article{hendrycks2023overview,
  title={An overview of catastrophic ai risks},
  author={Hendrycks, Dan and Mazeika, Mantas and Woodside, Thomas},
  journal={arXiv preprint arXiv:2306.12001},
  year={2023}
}

@article{olah2020zoom,
  title={Zoom in: An introduction to circuits},
  author={Olah, Chris and Cammarata, Nick and Schubert, Ludwig and Goh, Gabriel and Petrov, Michael and Carter, Shan},
  journal={Distill},
  volume={5},
  number={3},
  pages={e00024--001},
  year={2020}
}

@misc{rajamanoharan2024improving,
      title={Improving Dictionary Learning with Gated Sparse Autoencoders}, 
      author={Senthooran Rajamanoharan and Arthur Conmy and Lewis Smith and Tom Lieberum and Vikrant Varma and János Kramár and Rohin Shah and Neel Nanda},
      year={2024},
      eprint={2404.16014},
      archivePrefix={arXiv},
      primaryClass={cs.LG},
      url={https://arxiv.org/abs/2404.16014}, 
}

@article{bolukbasi2021interpretability,
  title={An interpretability illusion for bert},
  author={Bolukbasi, Tolga and Pearce, Adam and Yuan, Ann and Coenen, Andy and Reif, Emily and Vi{\'e}gas, Fernanda and Wattenberg, Martin},
  journal={arXiv preprint arXiv:2104.07143},
  year={2021}
}

@article{meng2022locating,
  title={Locating and editing factual associations in GPT},
  author={Meng, Kevin and Bau, David and Andonian, Alex and Belinkov, Yonatan},
  journal={Advances in Neural Information Processing Systems},
  volume={35},
  pages={17359--17372},
  year={2022}
}

@article{syed2023attribution,
  title={Attribution patching outperforms automated circuit discovery},
  author={Syed, Aaquib and Rager, Can and Conmy, Arthur},
  journal={arXiv preprint arXiv:2310.10348},
  year={2023}
}

@article{gao2024scaling,
  title={Scaling and evaluating sparse autoencoders},
  author={Gao, Leo and la Tour, Tom Dupr{\'e} and Tillman, Henk and Goh, Gabriel and Troll, Rajan and Radford, Alec and Sutskever, Ilya and Leike, Jan and Wu, Jeffrey},
  journal={arXiv preprint arXiv:2406.04093},
  year={2024}
}

@misc{lieberum2024gemma,
      title={Gemma Scope: Open Sparse Autoencoders Everywhere All At Once on Gemma 2}, 
      author={Tom Lieberum and Senthooran Rajamanoharan and Arthur Conmy and Lewis Smith and Nicolas Sonnerat and Vikrant Varma and János Kramár and Anca Dragan and Rohin Shah and Neel Nanda},
      year={2024},
      eprint={2408.05147},
      archivePrefix={arXiv},
      primaryClass={cs.LG},
      url={https://arxiv.org/abs/2408.05147}, 
}

@misc{bhalla2024interpreting,
      title={Interpreting CLIP with Sparse Linear Concept Embeddings (SpLiCE)}, 
      author={Usha Bhalla and Alex Oesterling and Suraj Srinivas and Flavio P. Calmon and Himabindu Lakkaraju},
      year={2024},
      eprint={2402.10376},
      archivePrefix={arXiv},
      primaryClass={cs.LG},
      url={https://arxiv.org/abs/2402.10376}, 
}

@misc{zhao2024steeringknowledgeselectionbehaviours,
      title={Steering Knowledge Selection Behaviours in LLMs via SAE-Based Representation Engineering}, 
      author={Yu Zhao and Alessio Devoto and Giwon Hong and Xiaotang Du and Aryo Pradipta Gema and Hongru Wang and Xuanli He and Kam-Fai Wong and Pasquale Minervini},
      year={2024},
      eprint={2410.15999},
      archivePrefix={arXiv},
      primaryClass={cs.CL},
      url={https://arxiv.org/abs/2410.15999}, 
}
\bibliographystyle{acl_natbib}

\newpage
\appendix
\onecolumn
\section{Appendix}

\subsection{Details on Pooling Methods}
\label{app:pooling_methods}
\subsubsection*{Top-\(N\) Feature Selection per Token}

In our approach, the top-\(N\) feature selection per token is performed as follows:

\paragraph{Step 1} For each token \(t\) in a sequence, we consider its corresponding SAE activation vector:
\[
f_t \in \mathbb{R}^m, \quad t=1,2,\dots,n,
\]
where \(m\) is the SAE dimension and \(n\) is the sequence length.

\paragraph{Step 2} For each token-level activation vector \(f_t\), we keep only the top \(N\) largest activation values, setting all other activations to zero:
\[
\tilde{f}_t[i] =
\begin{cases}
f_t[i], & \text{if } f_t[i]\text{ is among the top }N\text{ values in }f_t,\\[1mm]
0,      & \text{otherwise.}
\end{cases}
\]

\paragraph{Step 3} We then aggregate these sparse vectors across all tokens by summation to obtain a fixed-size sequence-level representation \(F\):
\[
F = \sum_{t=1}^{n} \tilde{f}_t.
\]

Thus, the selection is performed \emph{per token} independently (not across the entire dataset at once). This ensures each token contributes its most salient features, and then we aggregate token-level sparse activations into a sequence-level vector.

\subsubsection*{Top-\(N\) Mean-Difference Selection}

The top-\(N\) mean-difference selection method is a supervised feature selection approach performed at the dataset level, as follows:

\paragraph{Step 1} For each SAE dimension \(i\), compute the absolute difference between the mean activation for the positive class \(C^+\) and the negative class \(C^-\) over the entire training set:
\[
d_i \;=\;
\left|
\frac{1}{\lvert C^+\rvert} \sum_{x \in C^+} F_x(i)
\;-\;
\frac{1}{\lvert C^-\rvert} \sum_{x \in C^-} F_x(i)
\right|,
\]
where \(F_x(i)\) is the aggregated activation of dimension \(i\) for instance \(x\).

\paragraph{Step 2} Select the top \(N\) SAE dimensions with the largest \(d_i\) values. This selection is done once at the dataset level using the training data.

\paragraph{Step 3} For subsequent classification, keep only these top \(N\) dimensions for all instances.

In other words, the mean-difference selection is computed using activations aggregated across all tokens and all instances in the training dataset to identify globally discriminative SAE dimensions.

\subsection{Models and Dataset Information}
\label{app:model_info}

Table~\ref{tab:model_config} describes the configurations of the Gemma 2 models under study, including which layers are analyzed, the width of our SAE, and whether the model is base or instruction-tuned. 
These particular layers were selected based on availability of SAE widths across model sizes, and to reflect progression throughout the model.

\begin{table*}[ht]
    \centering
    \caption{Model Configurations and SAE Specifications. We analyze select intermediate layers (see \emph{Layers Analyzed}) to extract representations for the Stacked Autoencoder, whose width is indicated.}
    \label{tab:model_config}
    \begin{tabular}{llll}
        \toprule
        \textbf{Model} & \textbf{Layers Analyzed} & \textbf{SAE Width} & \textbf{Model Type} \\
        \midrule
        Gemma 2 2B & 5, 12, 19 & $2^{14}$, $2^{16}$ & Base \\
        Gemma 2 9B & 9, 20, 31 & $2^{14}$, $2^{17}$ & Base \\
        Gemma 2 9B-IT & 9, 20, 31 & $2^{14}$, $2^{17}$ & Instruction-tuned \\
        \bottomrule
    \end{tabular}
\end{table*}

Table~\ref{tab:dataset_specs} outlines each dataset used, specifying the type of task, a brief description, and the corresponding number of classes. 
These datasets focus on safety based tasks such as toxicity detection, and the multimodal datasets use the vision task such as CIFAR-100. 
Our goal was to test each model’s robustness across both domain (language vs.\ vision) and complexity (binary vs.\ multi-class classification), thereby evaluating classifiers applicability.

\begin{table}[ht]
    \centering
    \caption{Dataset Specifications, Task Descriptions, and Class Information. Each dataset is evaluated based on its primary task and class distribution. V) noted for vision tasks otherwise are pure text classification tasks}
    \label{tab:dataset_specs}
    \begin{tabular}{lp{5cm}c}
        \toprule
        \textbf{Dataset} & \textbf{Description} & \textbf{Classes} \\
        \midrule
        Multilingual Toxicity \cite{dementieva2024overview} 
            & Cross-lingual toxicity detection & 2 \\
        Election Questions \cite{anthropic_election_questions} 
            & Classify election-related claims  & 2 \\
        Reject Prompts \cite{arditi2024refusal} 
            & Detect unsafe instructions & 2 \\
        Jailbreak Classification \cite{jackhhao_jailbreak_classification} 
            & Detect model jailbreak attempts & 2 \\
        MASSIVE Intent \cite{fitzgerald2023massive} 
            & Massive intent classification & 60 \\
        MASSIVE Scenario \cite{fitzgerald2023massive} 
            & Massive scenario classification & 18 \\
        Banking77 \cite{Casanueva2020} 
            & Banking-related queries intent classification & 77 \\
        TweetEval Stance Abortion \cite{setfit_tweet_eval_stance_abortion} 
            & Stances on abortion: favor, against, neutral  & 3 \\
        NQ-Swap-original \cite{longpre2022entitybasedknowledgeconflictsquestion} 
            & Robustness testing with correct or incorrect factual information swapped QA & 2 \\
        V) CIFAR-100 \cite{krizhevsky2009learning} 
            & General image classification & 100 \\
        V) Oxford Flowers \cite{nelorth_oxford_flowers} 
            & Classification of 102 flower types & 102 \\
        V) Indian Food Images \cite{rajistics_indian_food_images} 
            & Classification of Indian dishes & 20 \\
        \bottomrule
    \end{tabular}
\end{table}

\FloatBarrier

\FloatBarrier
\newpage
\subsection{Performance variation on the width}
\label{app:width}
\begin{figure}[h!]
    \centering
    \includegraphics[width=\linewidth]{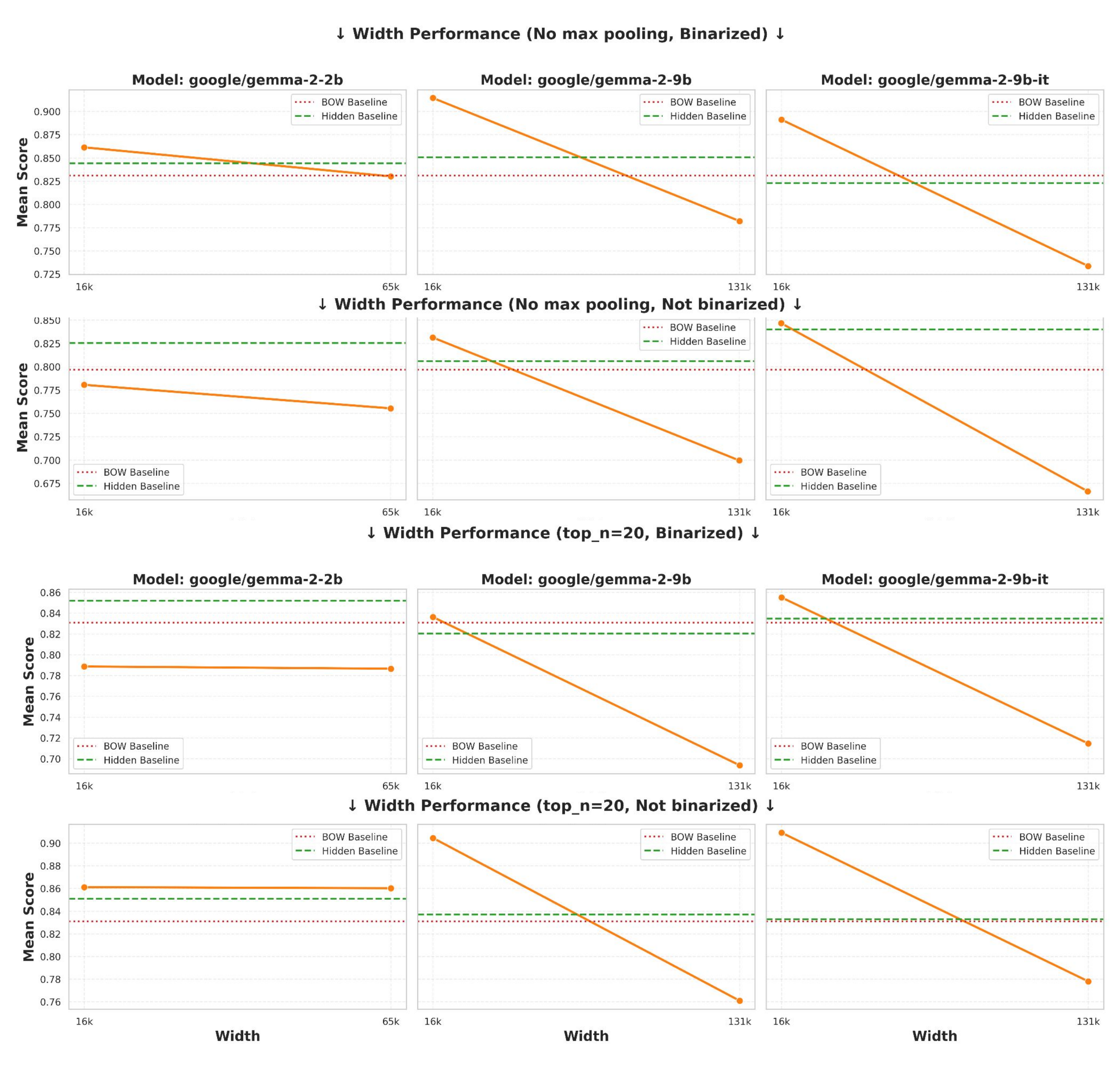}
    \caption{Performance evaluation of SAE feature transfer across different model widths for Gemma-2 models. Results are presented under different binarization and pooling settings, demonstrating a decline in mean score as width increases. The observed trends indicate that larger widths may reduce feature discriminability, particularly in non-binarized settings.}
    \label{fig:width-performance}
\end{figure}

We conduct an analysis of the effect of width scaling on full SAE features among our safety text classification tasks. The evaluation compares models with and without max pooling, as well as binarized and non-binarized activations, to determine their impact on classification performance. Consistently increasing the width results in decline in the mean score across all configurations, with the steepest drop observed in non-binarized cases, which is surprisingly different from \citet{sae_probing} demonstrate the opposite using mean-diff feature SAE selection. A complete table of our results across variations are available at the following anonymous link \url{https://docs.google.com/spreadsheets/d/1zUTXBdsorzthBLwMUoXNBP-X5lrUysnNL0iLYdBZ1HU/edit?usp=sharing}.

\newpage
\subsection{Multimodal performance}
\label{app:multimodal}

\begin{figure}[h!]
    \centering
    \includegraphics[width=\linewidth]{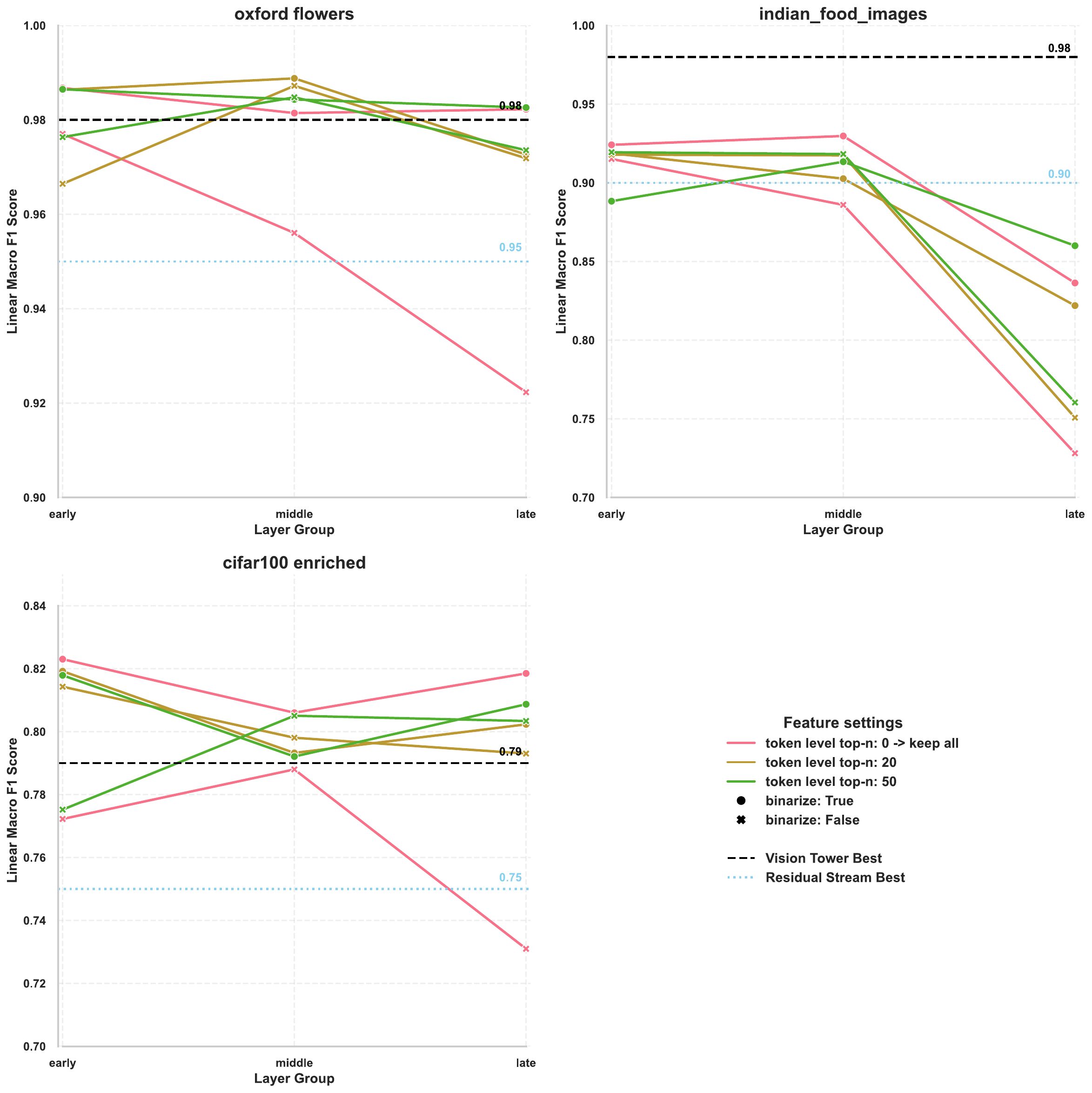}
    \caption{Performance of SAE features from gemmascope being utilised on activations derived from Peligemma 2 models. Token-n = 0 and binarization yielded overall best performance. These results also demonstrate the promise on direct SAE transfer in multimodal settings.}
    \label{fig:enter-label}
\end{figure}

We also implemented an unsupervised approach and analyzed the retrieved features to evaluate whether meaningful features could be identified through this transfer method among other models and pretrained SAEs. Initially, features were cleaned to remove those overrepresented across instances, which could add noise or reduce interpretability. 

Considering the CIFAR-100 dataset again, which comprises 100 labels with 100 instances per label, the expected maximum occurrence of any feature under uniform distribution is approximately 100. To address potential anomalies, a higher threshold of 1000 occurrences was selected as the cutoff for identifying and excluding overrepresented features. This conservative threshold ensured that dominant, potentially less informative features were removed while retaining those likely to contribute meaningfully to the analysis.

In this study, we also tried the Intel Gemma-2B LLaVA 1.5-based model (Intel/llava-gemma-2b) \cite{hinck2024llavagemma} as the foundation for our experiments. For feature extraction, we incorporate pre-trained SAEs from jbloom/Gemma-2b-Residual-Stream-SAEs (RELU-based), trained on the Gemma-1-2B model. These SAEs include 16,384 features (an expansion factor of 8 × 2048) and are designed to capture sparse and interpretable representations.

After cleaning, we examined the retrieved features across different model layers (0–12 of 19 layers). We found that deeper layers exhibited increasingly useful/relevant features.

Below, we provide some examples of retrieved features from both high-performing and underperforming classes, demonstrating the range of interpretability outcomes.

\subsection{Top retrieved features}
\label{app:multimodal_retrieve}

\begin{longtable}{|c|c|p{7cm}|}
\hline
\textbf{Category} & \textbf{Layer} & \textbf{Top 2 Features (Occurrences)} \\
\hline
\endfirsthead

\multicolumn{3}{c}%
{{\bfseries -- Continued from previous page --}} \\
\hline
\textbf{Category} & \textbf{Layer} & \textbf{Top 2 Features (Occurrences)} \\
\hline
\endhead

\hline \multicolumn{3}{|r|}{{Continued on next page}} \\ \hline
\endfoot

\hline
\endlastfoot

Dolphin & Layer 0 & Technical information related to cooking recipes and server deployment (30/100) \\
 &  & References to international topics or content (26/100) \\
\hline
Dolphin & Layer 6 & Phrases related to a specific book title: \textit{The Blue Zones} (25/100) \\
 &  & Mentions of water-related activities and resources in a community context (17/100) \\
\hline
Dolphin & Layer 10 & Terms related to underwater animals and marine research (88/100) \\
 &  & Actions involving immersion, dipping, or submerging in water (61/100) \\
\hline
Dolphin & Layer 12 & Terms related to oceanic fauna and their habitats (77/100) \\
 &  & References to the ocean (53/100) \\
\hline
Dolphin & Layer 12-it & Mentions of the ocean (60/100) \\
 &  & Terms related to maritime activities, such as ships, sea, and naval battles (40/100) \\
\hline
Skyscraper & Layer 0 & Information related to real estate listings and office spaces (11/100) \\
 &  & References to sports teams and community organizations (7/100) \\
\hline
Skyscraper & Layer 6 & Details related to magnification and inspection, especially for physical objects and images (32/100) \\
 &  & Especially for physical objects and images (28/100) \\
\hline
Skyscraper & Layer 10 & References to physical structures or buildings (68/100) \\
 &  & Character names and references to narrative elements in storytelling (62/100) \\
\hline
Skyscraper & Layer 12 & References to buildings and structures (87/100) \\
 &  & Locations and facilities related to sports and recreation (61/100) \\
\hline
Skyscraper & Layer 12-it & Terms related to architecture and specific buildings (78/100) \\
 &  & References to the sun (57/100) \\
\hline
Boy & Layer 0 & References to sports teams and community organizations (17/100) \\
 &  & Words related to communication and sharing of information (10/100) \\
\hline
Boy  & Layer 6 & Phrases related to interior design elements, specifically focusing on color and furnishings (52/100) \\
 &  & Hair styling instructions and descriptions (25/100) \\
\hline
Boy  & Layer 10 & Descriptions of attire related to cultural or traditional clothing (87/100) \\
 &  & References to familial relationships, particularly focusing on children and parenting (83/100) \\
\hline
Boy  & Layer 12 & Words associated with clothing and apparel products (89/100) \\
 &  & Phrases related to parental guidance and involvement (60/100) \\
\hline
Boy  & Layer 12-it & Patterns related to monitoring and parental care (88/100) \\
 &  & Descriptions related to political issues and personal beliefs (67/100) \\
\hline
Cloud & Layer 0 & Possessive pronouns referring to one's own or someone else's belongings or relationships (4/100) \\
 &  & References to sports teams and community organizations (3/100) \\
\hline
Cloud & Layer 6 & Descriptive words related to weather conditions (24/100) \\
 &  & Mentions of astronomical events and celestial bodies (21/100) \\
\hline
Cloud & Layer 10 & Terms related to aerial activities and operations (62/100) \\
 &  & References and descriptions of skin aging or skin conditions (59/100) \\
\hline
Cloud & Layer 12 & Themes related to divine creation and celestial glory (92/100) \\
 &  & Terms related to cloud computing and infrastructure (89/100) \\
\hline
Cloud & Layer 12-it & The word "cloud" in various contexts (80/100) \\
 &  & References to the sun (47/100) \\
\hline

\end{longtable}

\FloatBarrier
\newpage

\subsection{Performance Tables}
\label{app:performance_table}

Below we present the full results for evaluating our multilingual toxicity classification experiments, focusing on different feature extraction methods, top-$n$ feature selection, and the overall experimental design. 

\begin{table}[ht]
\centering
\caption{Multilingual Toxicity Classification Performance Comparison}
\label{tab:results}
\begin{tabular}{l l c c c c c c}
\toprule
Model & Transfer & \multicolumn{3}{c}{SAE Features} & \multicolumn{3}{c}{Hidden States} \\
 & & Layer 9 & 20 & 31 & Layer 9 & 20 & 31 \\
\midrule
Gemma2 - 9B & Original   & 0.759 & \textbf{0.794} & 0.766 & 0.772 & 0.792 & 0.765 \\
 & Translated              & 0.763 & \textbf{0.798} & 0.771 & 0.771 & 0.794 & 0.766 \\
Gemma2 - 9B IT & Original & 0.754 & \textbf{0.784} & 0.751 & 0.755 & 0.770 & 0.755 \\
 & Translated              & 0.761 & \textbf{0.778} & 0.753 & 0.761 & 0.776 & 0.747 \\
\bottomrule
\end{tabular}
\end{table}

\paragraph{SAE Features vs.\ Hidden States.}
Table~\ref{tab:results} reports macro F1 scores for two Gemma2 9B model variants (base and instruction-tuned), comparing:
\begin{enumerate}
    \item \textit{SAE Features:} Representations learned by a Sparse Autoencoder at specific layers.
    \item \textit{Hidden States:} Direct residual stream outputs/hidden states from the same transformer layers.
\end{enumerate}
We evaluate both \emph{Original} (multilingual) and \emph{Translated} (All translated to English) test sets. 
Across most settings, SAE-based features at layer 20 or 31 produce competitive (often superior) results, suggesting that deeper layers encode richer semantic information for toxicity detection. 
The instruction-tuned model (Gemma2 - 9B IT) also benefits from SAE features, although its absolute scores are slightly lower than the base model’s best results, surprisingly, on both using full SAE features and hidden states.

\begin{table}[ht]
\centering
\caption{Comparison of F1 scores across different layers and top-$N$ token selections. 
\textbf{top-$N$} indicates evaluation on the top 10, 20, or 50 \textbf{mean top-diff SAE features}. 
\textbf{Original} refers to the original input language, while \textbf{Translated} corresponds to translated input to English. 
Bold values highlight the highest scores for each row.}
\label{tab:layer_eval}
\vspace{1em}
\begin{tabular}{l l ccc ccc ccc}
\toprule
 & Transfer & \multicolumn{3}{c}{Top 10} & \multicolumn{3}{c}{Top 20} & \multicolumn{3}{c}{Top 50} \\
\cmidrule(lr){3-5} \cmidrule(lr){6-8} \cmidrule(lr){9-11}
\textbf{Model} & \textbf{Setting} & L9 & L20 & L31 & L9 & L20 & L31 & L9 & L20 & L31 \\
\midrule
\multirow{2}{*}{Gemma2 - 9B}    & Original     & 0.72 & 0.76 & \textbf{0.79}  & 0.72 & 0.76 & \textbf{0.79}  & 0.72 & 0.76 & \textbf{0.79} \\
                                & Translated   & 0.72 & 0.76 & \textbf{0.78}  & 0.72 & 0.76 & \textbf{0.78}  & 0.72 & 0.76 & \textbf{0.78} \\
\midrule
\multirow{2}{*}{Gemma2 - 9B IT} & Original     & 0.72 & 0.74 & \textbf{0.77}  & 0.72 & 0.74 & \textbf{0.77}  & 0.72 & 0.74 & \textbf{0.77} \\
                                & Translated   & 0.72 & 0.73 & \textbf{0.76}  & 0.72 & 0.73 & \textbf{0.76}  & 0.72 & 0.73 & \textbf{0.76} \\
\bottomrule
\end{tabular}
\end{table}

In the table above, we investigate selecting only the top 10, 20, or 50 most salient SAE features. Interestingly, reduced features can maintain or sometimes even slightly improve macro F1 performance.


\FloatBarrier

\subsection{Cross Lingual Transfer of Feature Activations}
\label{app:cross_lingual_features}
\begin{figure}[ht]
    \centering
    \includegraphics[width=\linewidth]{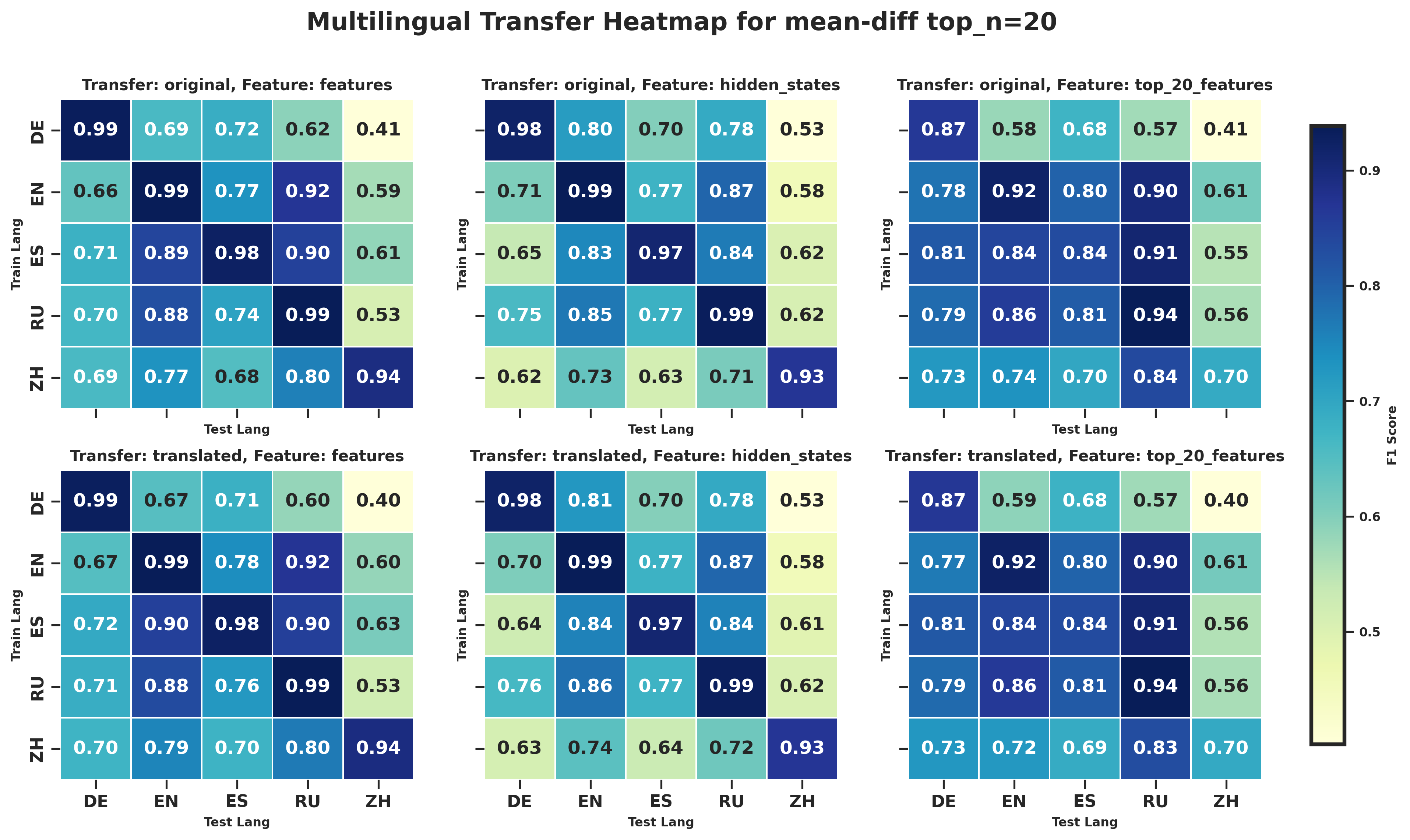}
    \caption{Average F1 scores for each training language (y-axis) versus test language (x-axis). 
    We compare hidden states, SAE features, and top-$n$ feature selection. 
    The top row shows models trained on native-language datasets; 
    the bottom row uses English-translated data for training. 
    Darker cells indicate higher F1 performance.}
    \label{fig:multilingual_heatmap}
\end{figure}

\begin{figure}
    \centering
    \includegraphics[width=\linewidth]{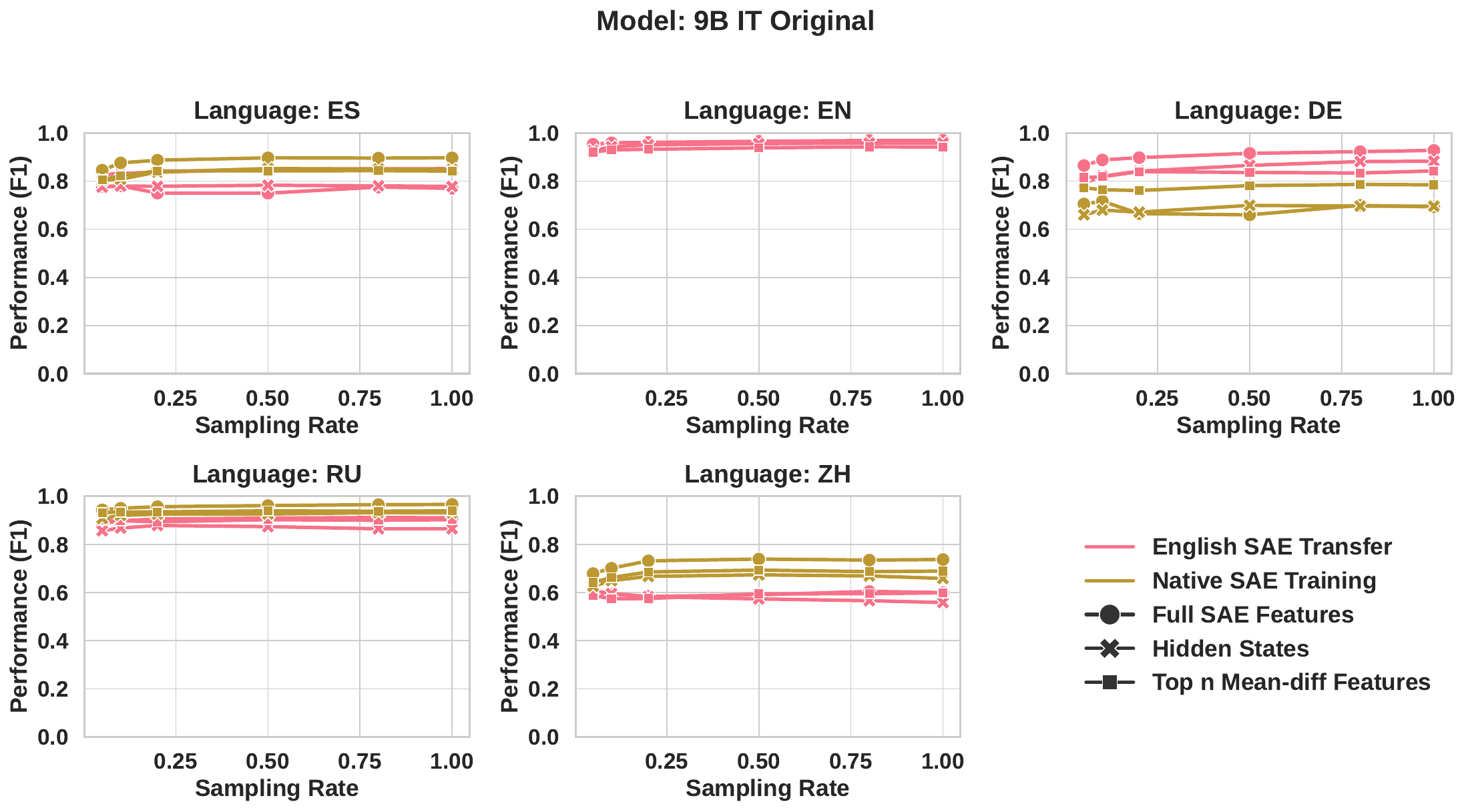}
    \caption{Performance vs.\ sampling rate for the 9B \emph{instruction-tuned} model on \emph{original-language} data. 
    The x-axis is the sampling rate (from 0.25 to 1.0), and the y-axis is F1 score. 
    Each subplot corresponds to a different language (ES, DE, EN, RU, ZH), while line colors distinguish \emph{Native SAE Training} from \emph{English SAE Transfer}. 
    Markers reflect the feature extraction approach (\emph{features}, \emph{hidden\_states}, or \textbf{mean difference} \emph{top\_n\_features}).}
    \label{fig:lineplot_9b_it_original}
\end{figure}

\begin{figure}
    \centering
    \includegraphics[width=\linewidth]{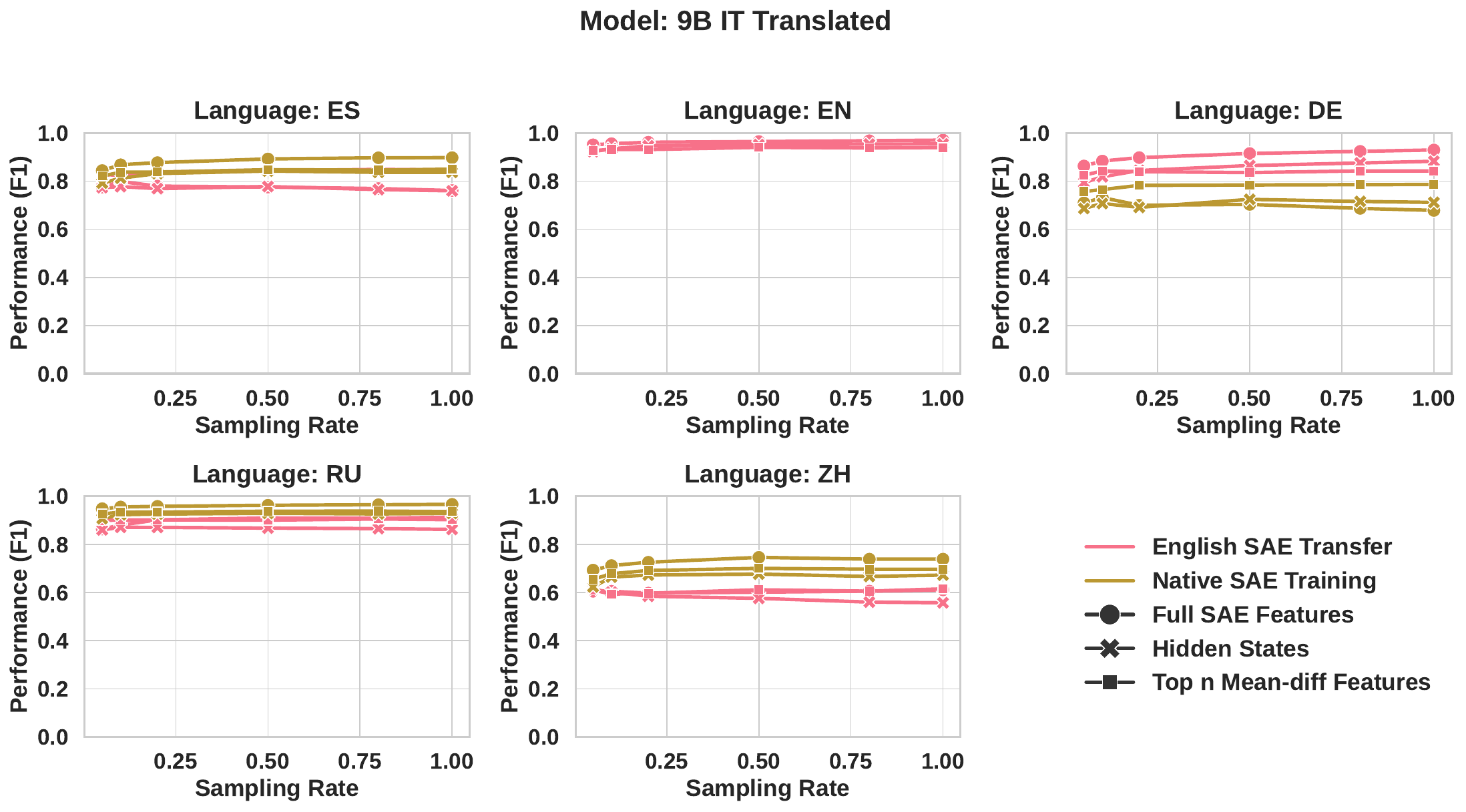}
    \caption{Performance vs.\ sampling rate for the 9B \emph{instruction-tuned} model on different \emph{translated-language} data. 
    As in Figure~\ref{fig:lineplot_9b_it_original}, the x-axis shows sampling rate, the y-axis is F1, and subplots detail performance across ES, DE, EN, RU, and ZH. 
    Lines and markers compare \emph{English SAE Transfer} to \emph{Native SAE Training} under different feature types.}
    \label{fig:lineplot_9b_it_translated}
\end{figure}

\begin{figure}
    \centering
    \includegraphics[width=\linewidth] {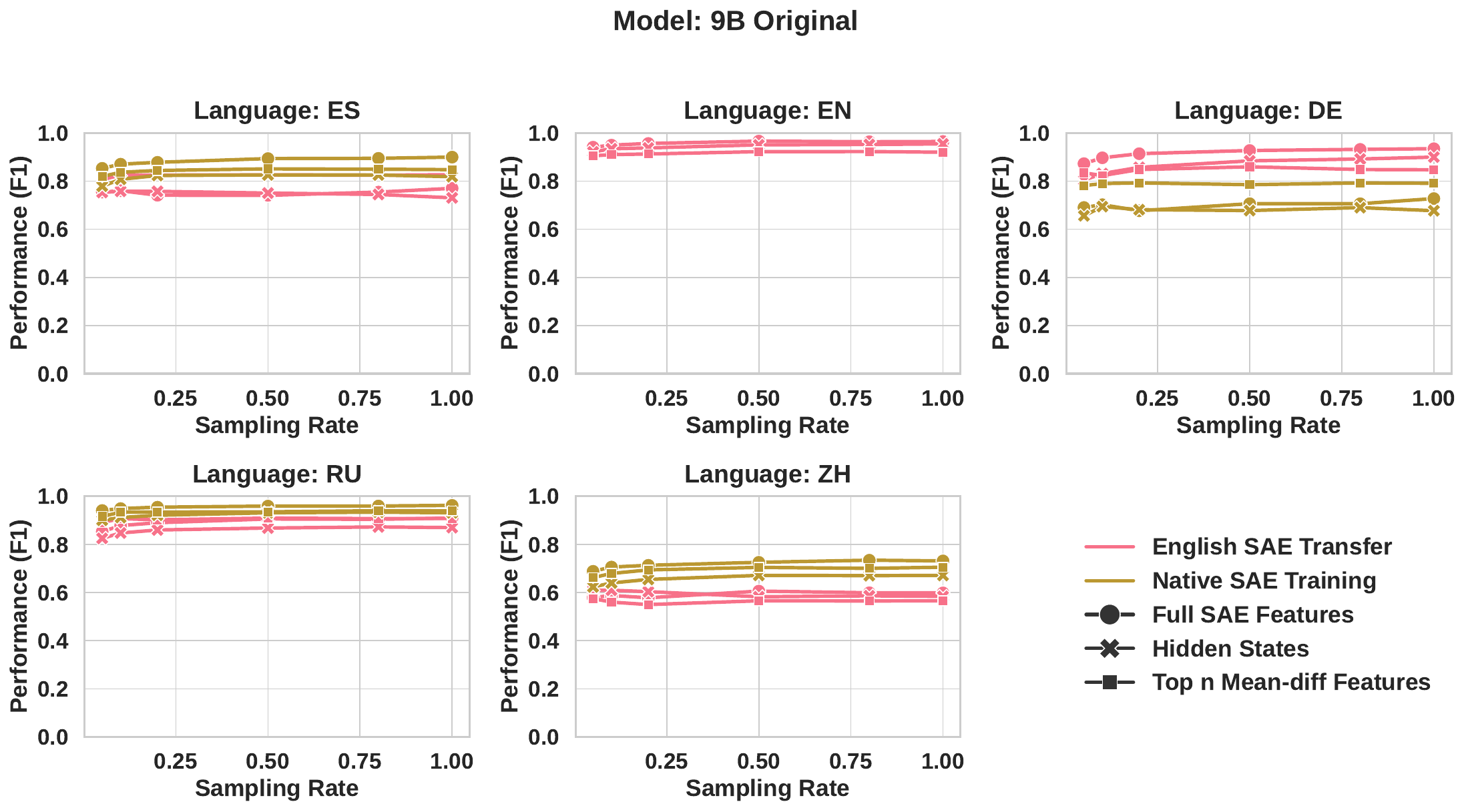}
    \caption{Performance vs.\ sampling rate for the 9B \emph{base} model using \emph{original-language} data. 
    Subplots again separate ES, DE, EN, RU, and ZH. 
    The curves illustrate how training type (Native vs.\ English transfer) and feature extraction (full features, hidden states, \textbf{mean difference} top $n$ features) affect F1 across varying sampling rates.}
    \label{fig:lineplot_9b_original}
\end{figure}

\begin{figure}
    \centering
    \includegraphics[width=\linewidth]{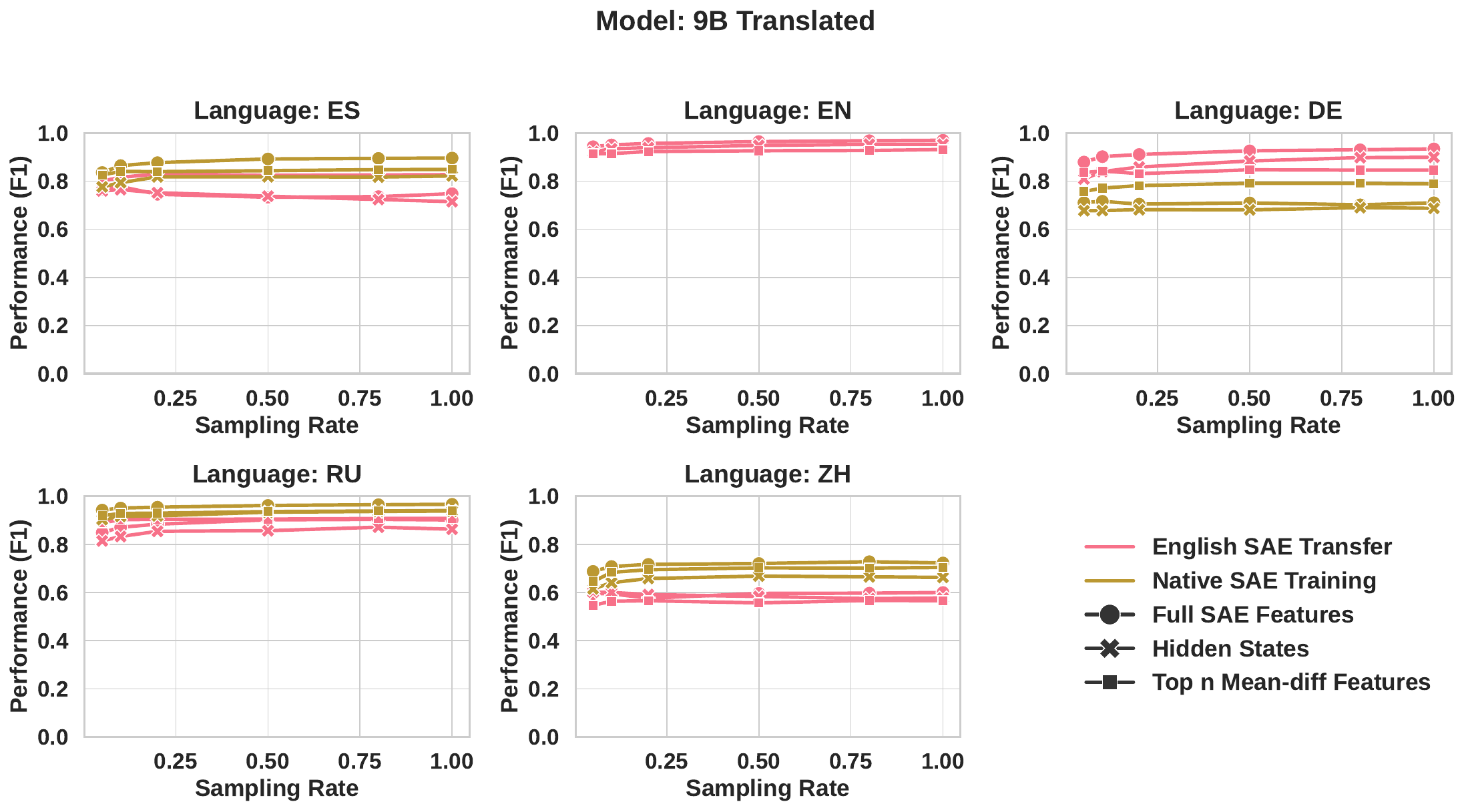}
    \caption{Performance vs.\ sampling rate for the 9B \emph{base} model using \emph{translated} datasets. 
    The x-axis is sampling rate, the y-axis is F1, and each subplot is a distinct target language. 
    Color and marker styles reflect training type and feature extraction, as in prior figures.}
    \label{fig:lineplot_9b_translated}
\end{figure}

A more detailed set of visualizations are provided below showing how feature extraction methods perform when transferring across different languages. We first show a high-level summary of cross-lingual transfer via a heatmap (Figure~\ref{fig:multilingual_heatmap}), then we provide a series of line plots (Figures~\ref{fig:lineplot_9b_it_original}--\ref{fig:lineplot_9b_translated}) illustrating performance versus sampling rate for five target languages. These plots compare \emph{Native SAE Training} with \emph{English SAE Transfer} under three feature extraction strategies: \emph{full SAE features}, \emph{hidden states}, and \textbf{mean difference} \emph{top-$n$ SAE features}.

\paragraph{Analysis of Feature Overlap.}
Table~\ref{tab:scores_overlap} compares F1 scores for different training approaches (\emph{English Transfer}, \emph{Native}, and \emph{Translated SAE}) across five languages (DE, EN, ES, RU, ZH). The \emph{Overlap} columns indicate how many of the top 20 SAE features are shared with each respective training scheme. As expected, each model has a complete overlap (1.000) with its own native features. In contrast, cross-lingual overlaps (e.g., \emph{Overlap English} for Spanish or Chinese) are comparatively low (often around 0.06--0.26). Top 20 features were stored for each model trained on a language. Overlaps were calculated as standard jaccard similarities measures between train and test language sites, where we compare the features from the training set of one language to that of the top 20 features derived during training on the test language. For example, English-Spanish overlap is calculated using the top 20 SAE features derived from logistic regression training on the English dataset, and the top 20 features derived from logistic regression training on the Spanish Dataset. We then compute the similarity metric between the two.

\begin{table*}[ht]
    \centering
    \caption{F1 Scores and Overlap for Models and Test Languages. 
    F1 scores are reported for three evaluation strategies: 
    \textbf{F1 (EN-T)}: Trained on English SAE features and tested on other languages (Transfer), 
    \textbf{F1 (N)}: Trained and tested natively, 
    \textbf{F1 (Tr-SAE)}: Trained on translated inputs with extracted SAE features. 
    Overlap measures indicate representation similarity: 
    \textbf{Ovlp (EN)}: Overlap with English Transfer, 
    \textbf{Ovlp (Tr)}: Overlap with Translated SAE.}
    \label{tab:scores_overlap}
    \begin{tabular}{llccc cc}
        \toprule
        \textbf{Model} & \textbf{Lang} & \textbf{F1 (EN-T)} & \textbf{F1 (N)} & \textbf{F1 (Tr-SAE)} & \textbf{Ovlp (EN)} & \textbf{Ovlp (Tr)} \\
        \midrule
        9b    & DE  & 0.710 & 0.945 & 0.708 & 0.098 & 0.099 \\
        9b    & EN  & --    & 0.969 & --    & --    & --    \\
        9b    & ES  & 0.768 & 0.926 & 0.771 & 0.212 & 0.200 \\
        9b    & RU  & 0.888 & 0.973 & 0.886 & 0.237 & 0.221 \\
        9b    & ZH  & 0.592 & 0.856 & 0.593 & 0.061 & 0.064 \\
        \cmidrule(lr){1-7}
        9b it & DE  & 0.722 & 0.941 & 0.723 & 0.093 & 0.089 \\
        9b it & EN  & --    & 0.969 & --    & --    & --    \\
        9b it & ES  & 0.792 & 0.928 & 0.790 & 0.207 & 0.209 \\
        9b it & RU  & 0.903 & 0.973 & 0.903 & 0.263 & 0.253 \\
        9b it & ZH  & 0.599 & 0.858 & 0.602 & 0.086 & 0.071 \\
        \bottomrule
    \end{tabular}
\end{table*}

Despite relatively small overlaps in top features, the \emph{English Transfer} and \emph{Translated SAE} configurations can still yield competitive F1 scores (e.g., RU with English Transfer at 0.888 or 0.903 for instruction-tuned). This suggests that, although the top features in one language are not strictly identical to those in another, a significant subset of high-impact features appears useful across languages. At the same time, the strongest performance generally occurs under \emph{Native} training.

\subsubsection{Full SAE learns classifiers find different features than Mean-diff top-N features}

\begin{figure}[ht]
    \centering
    \includegraphics[width=\linewidth]{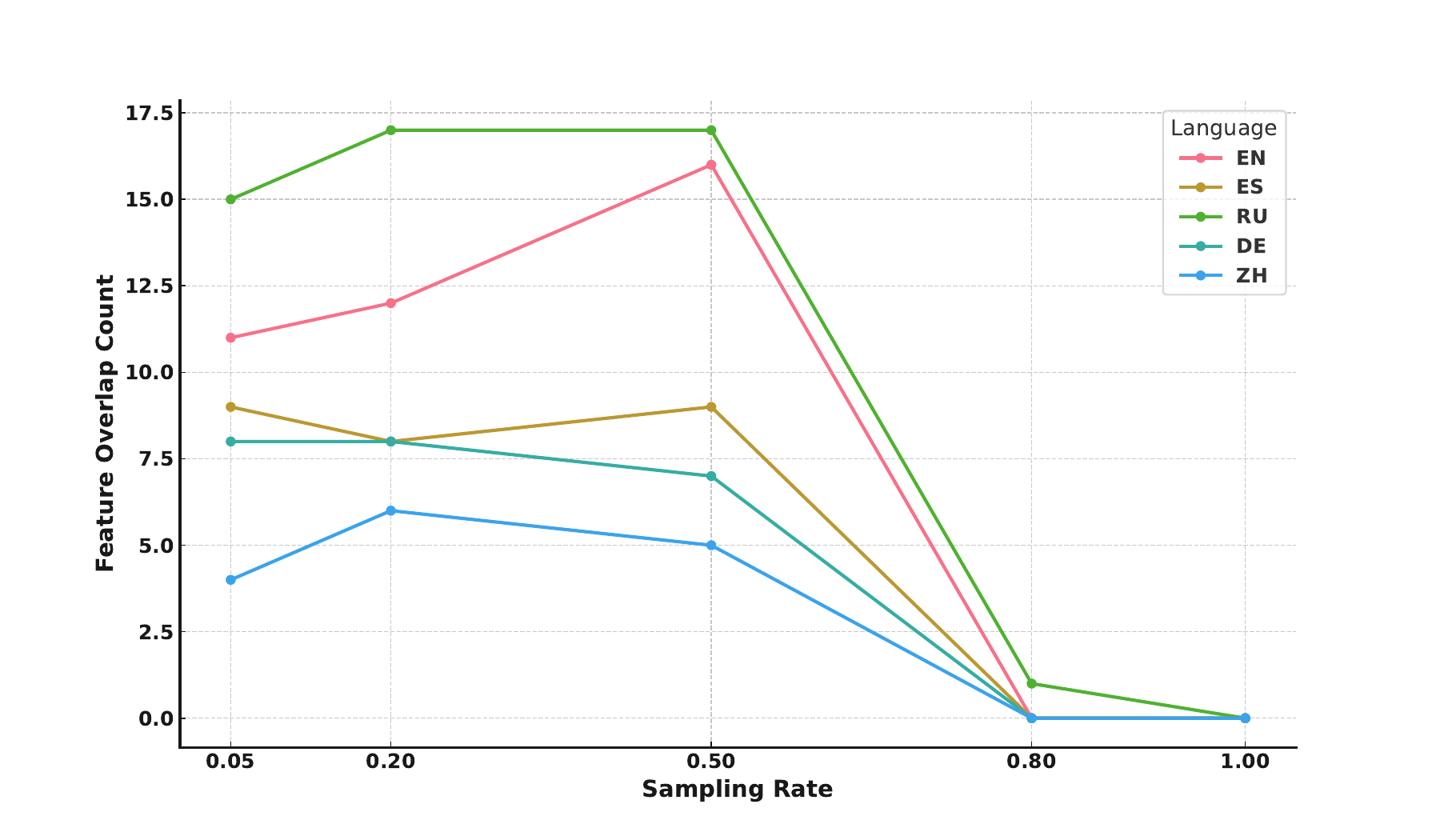}
    \caption{Feature overlap count between Full SAE Top-20 and Mean-Difference Top-20 feature selection across sampling rates among native-language trained SAEs (from 9B-IT, layer 31). Higher overlap suggests greater consistency in feature selection between the two methods.}
    \label{fig:feature_overlap}
\end{figure}

As we have seen in Figure 10-13, our Full SAE learns features outperform the Mean-diff Top-20 features. This makes sense because our features are learned through supervision, while the other method is done by naive clustering. You can also see that the top-20 "useful features" found by two different method from 9B-IT model is different in Figure \ref{fig:feature_overlap}. As we use more data, the overlap fully got washed out.

\FloatBarrier

\subsection{Action prediction}
\label{app:action_paired_plot}

\begin{figure}
    \centering
    \includegraphics[width=\linewidth]{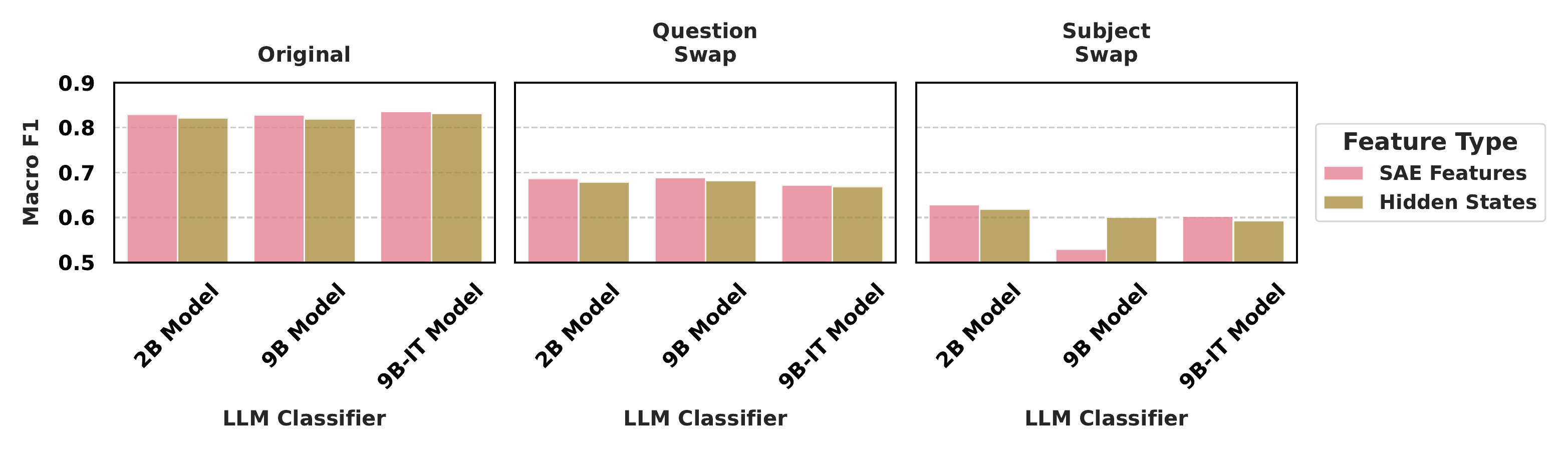}
    \caption{Paired bar plot for hidden state compared to SAE feature performance for behavior prediction across datasets.}
    \label{fig:action_paired}
\end{figure}

below we show the disaggregated performance of SAE features vs.\ hidden states ability to predict a model's actions or behaviors across multiple task scenarios. Specifically, we focus on the 9B instruction-tuned model (\emph{9b it}) under three dataset conditions:
\begin{enumerate}
    \item \textbf{Original questions without context:} Queries posed directly with no additional background.
    \item \textbf{Questions with correct context:} Queries augmented by relevant information aligned with the true scenario.
    \item \textbf{Questions with incorrect context:} Queries intentionally combined with misleading or contradictory statements.
\end{enumerate}

Figure~\ref{fig:action_paired} presents a paired bar plot that compares \emph{hidden states} (gold bars) and \emph{SAE features} (pink bars) for predicting whether the model will respond with a particular action or behavior. Each subplot corresponds to a different dataset, illustrating how these features perform under various context conditions. Notably, the SAE-based classifier often achieves performance levels on par with or superior to the raw hidden-state baseline, suggesting that SAE features may help isolate key aspects of the model’s decision-making process. This pattern holds across original questions (no context) as well as questions provided with correct or incorrect context, indicating that SAE features can enhance interpretability and robustness in action prediction tasks.

\subsection{Action Features}
\label{app:action_autointerp}

To further investigate how these learned representations drive action prediction, we highlight in the tables below the top classifier features for the original and no context scenario in the middle layer setting, reflecting the core layers from which features are extracted. 

The goal would be to identify if similar concepts are activated across model sizes e.g. are features from the 2b similar to the concepts on the 9b-it that is trying to predict its own behaviour? These tables help reveal whether similar conceptual features emerge across different context conditions (e.g., \textit{No Context} vs.\ \textit{original}) or whether the model learns context-specific indicators tied to the question setup.

\begin{table}[ht]
    \centering
    \caption{Feature Comparison for Dataset: No Context, Layer: middle}
    \label{table:No_Context_layer_middle}
    \resizebox{\textwidth}{!}{%
\begin{tabular}{lll}
\toprule
Feature (Model google/gemma-2-2b) & Feature (Model google/gemma-2-9b) & Feature (Model google/gemma-2-9b-it) \\
\midrule
10: terms related to programming languages & 11: terms related to competition and ranking & 319:  phrases that denote parts of a whole \\
444: phrases indicating a scarcity or lack of something & 3143: expressions of pride and accomplishments & 1513: phrases related to raising awareness and advocacy for various social issues \\
632:  car dealership and financing-related terminology & 4152: technical terms and concepts related to data streaming and manipulation & 2032:  topics related to societal norms and expectations \\
1373: conjunctive phrases that express relationships or connections between multiple elements & 4316: authenticity and sincerity in relationships and choices & 7597: references to publishers and publication details \\
4214: phrases relating to economic inequality and socio-political commentary & 4771: terms related to the emission of light and radiation in various contexts & 8568:  legal terminology and concepts related to administrative and tax liability \\
5593:  terms related to switching or transitions & 8741: instances of the verb "pass" and its variations in context & 9520: references to applications, their requirements, and the processes involved in their submission and approval \\
10177:  references to procedures and protocols & 9153: phrases related to approaching critical points or thresholds & 9912: elements and methods related to API request handling and asynchronous processing \\
10316: terms related to study design and data analysis methods & 12185:  references to sanctions and their implications & 12025: references to meetings and discussions \\
13181: phrases that refer to taking or maintaining control or responsibility & 13192: references to biblical imagery and themes related to prophecy and divine intervention & 13586: common phrases or templates in written dialogues \\
15360: periods at the end of sentences & 13510:  code-related terminology and concepts in programming languages & 14004:  occurrences of specific events and their frequency in a legal or conversational context \\
\bottomrule
\end{tabular}

    }
\end{table}

\begin{table}[ht]
    \centering
    \caption{Feature Comparison for Dataset: Original, Layer: middle}
    \label{table:Original_layer_middle}
    \resizebox{\textwidth}{!}{%
\begin{tabular}{lll}
\toprule
Feature (Model google/gemma-2-2b) & Feature (Model google/gemma-2-9b) & Feature (Model google/gemma-2-9b-it) \\
\midrule
1189: commands or instructions related to processing data or managing functions & 1976: technical terms and phrases related to experimental setups and measurements & 557: mentions of personal identity and name references \\
3563: syntax related to resource management and context management in programming (e.g., using "with" and "using" statements) & 4864:  cooking-related terms and ingredients & 1489: instances of dialogue and conversational exchanges \\
4705: numerical and alphanumeric sequences, likely related to coding or technical details & 5181:  components of code related to database operations and responses & 2297:  technical programming concepts and syntax elements \\
5382: phrases related to customer engagement and interactions in a business context & 6672: medical terminology related to women's health conditions & 3084:  contact information and email addresses \\
7360: elements related to function and method definitions & 6729:  mathematical symbols and notations & 4110: code structure and syntax elements in programming \\
10140: elements related to programming structures and their definitions & 7656:  punctuation and formatting markers typical in academic citations & 5465:  phrases related to legal and ethical violations \\
10421:  references to programming languages, libraries, and frameworks related to system and web development & 7926:  terms related to weights and their configurations in neural networks & 6645: references to mathematical variables and parameters associated with functions and their behaviors \\
12396:  assignment operations in code & 9384: terms related to exercise and physical activity & 7196:  references to upcoming events or competitions \\
13999: array declarations and manipulations in code & 9708: terms related to crime and legal issues & 9384:  proper nouns related to people, places, and institutions \\
14399: currency symbols and monetary values & 13547:  programming-related syntax and structure & 13338: words related to programming or software-related language components \\
\bottomrule
\end{tabular}

    }
\end{table}

\paragraph{High-Level Consistencies Across Models.}
Across the tables comparing 2B, 9B, and 9B-IT, we see frequent mentions of programming-related features (e.g., code syntax, function definitions, data structures). Such technical elements dominate many of the top features identified by our \textit{autointerpretable} definitions. However, we also observe several non-programming references (e.g., legal terminology, societal or economic concepts) shared across models—particularly at middle or late layers.

An example we observe is the presence of \emph{Economic and Socio-Political Commentary} across models. The 2B model identifies phrases relating to “economic inequality and socio-political commentary” (Feature 4214), whereas 9B-IT surfaces “legal terminology and concepts related to administrative and tax liability” (Feature 8568). Both target broader sociopolitical or legal contexts.

It is important to note that our similarity claims are constrained by the level of granularity in \textit{autointerpretable} annotations. Different feature IDs may describe related or overlapping real-world concepts, even if they are not labeled identically. At a high level, these tables suggest that all three Gemma-2 variants (2B, 9B, and 9B-IT) learn to capture similar domains, with broad thematic parallels (legal frameworks, social dynamics, etc.) emerging beyond mere code-based patterns. Thus, even though the precise feature names differ, it appears plausible that many of these salient features reflect similar underlying concepts.

\end{document}